\documentclass[runningheads]{llncs}
\usepackage{graphicx}
\usepackage{comment}
\usepackage{amsmath,amssymb}
\usepackage{pifont}
\usepackage{color}
\usepackage{url}
\usepackage{hyperref}
\usepackage{cleveref}
\usepackage{breakcites}
\usepackage{caption}
\usepackage{algorithm}
\usepackage{algpseudocode}
\newcommand{\miou}[1]	{\text{mIoU}_\text{#1}}
\newcommand{\smalleq}	{\!=\!}
\newcommand{\xmark} {\textbf{\ding{55}}}
\newcommand{\cmark} {\textbf{\checkmark}}
\newcommand{\none} {\multicolumn{1}{c|}{-}}
\newcommand{\noneB} {\multicolumn{1}{c}{-}}
\crefname{figure}{Fig.}{Figs.}
\crefname{section}{Section}{Sections}
\crefname{table}{Table}{Tables}
\crefname{algorithm}{Algorithm}{Algorithms}
\crefname{equation}{Equation}{Equations}
\crefname{appendix}{Appendix}{Appendix}

\begin{document}

\title{PARMESAN: Parameter-Free Memory Search and Transduction for Dense Prediction Tasks}
\titlerunning{PARMESAN}

\author{Philip Matthias Winter*, Maria Wimmer*, David Major, Dimitrios Lenis, Astrid Berg, Theresa Neubauer, Gaia Romana De Paolis, Johannes Novotny, Sophia Ulonska, Katja Bühler\thanks{This is the author's accepted manuscript of a paper published in Lecture Notes in Computer Science (LNCS), volume 15297, Proceedings of DAGM GCPR 2024, DOI: \url{https://doi.org/10.1007/978-3-031-85181-0_1}.}}
\authorrunning{P.M.~Winter, M.~Wimmer et al.}

\institute{VRVis GmbH, Donau-City-Straße 11, 1220 Vienna\\
\email{office@vrvis.at}\\
\url{https://www.vrvis.at}\\
*These authors contributed equally to this work.}

\maketitle

\begin{abstract}
This work addresses flexibility in deep learning by means of transductive reasoning. For adaptation to new data and tasks, e.g., in continual learning, existing methods typically involve tuning learnable parameters or complete re-training from scratch, rendering such approaches unflexible in practice. We argue that the notion of separating computation from memory by the means of transduction can act as a stepping stone for solving these issues. We therefore propose PARMESAN (\textbf{par}ameter-free \textbf{me}mory \textbf{s}earch \textbf{a}nd transductio\textbf{n}), a scalable method which leverages a memory module for solving dense prediction tasks. At inference, hidden representations in memory are being searched to find corresponding patterns. In contrast to other methods that rely on continuous training of learnable parameters, PARMESAN learns via memory consolidation simply by modifying stored contents. Our method is compatible with commonly used architectures and canonically transfers to 1D, 2D, and 3D grid-based data. The capabilities of our approach are demonstrated at the complex task of continual learning. PARMESAN learns by 3-4 orders of magnitude faster than established baselines while being on par in terms of predictive performance, hardware-efficiency, and knowledge retention.
\end{abstract}

\section{Introduction}

\begin{figure}[t]
	\centering
	\includegraphics[width=0.75\linewidth]{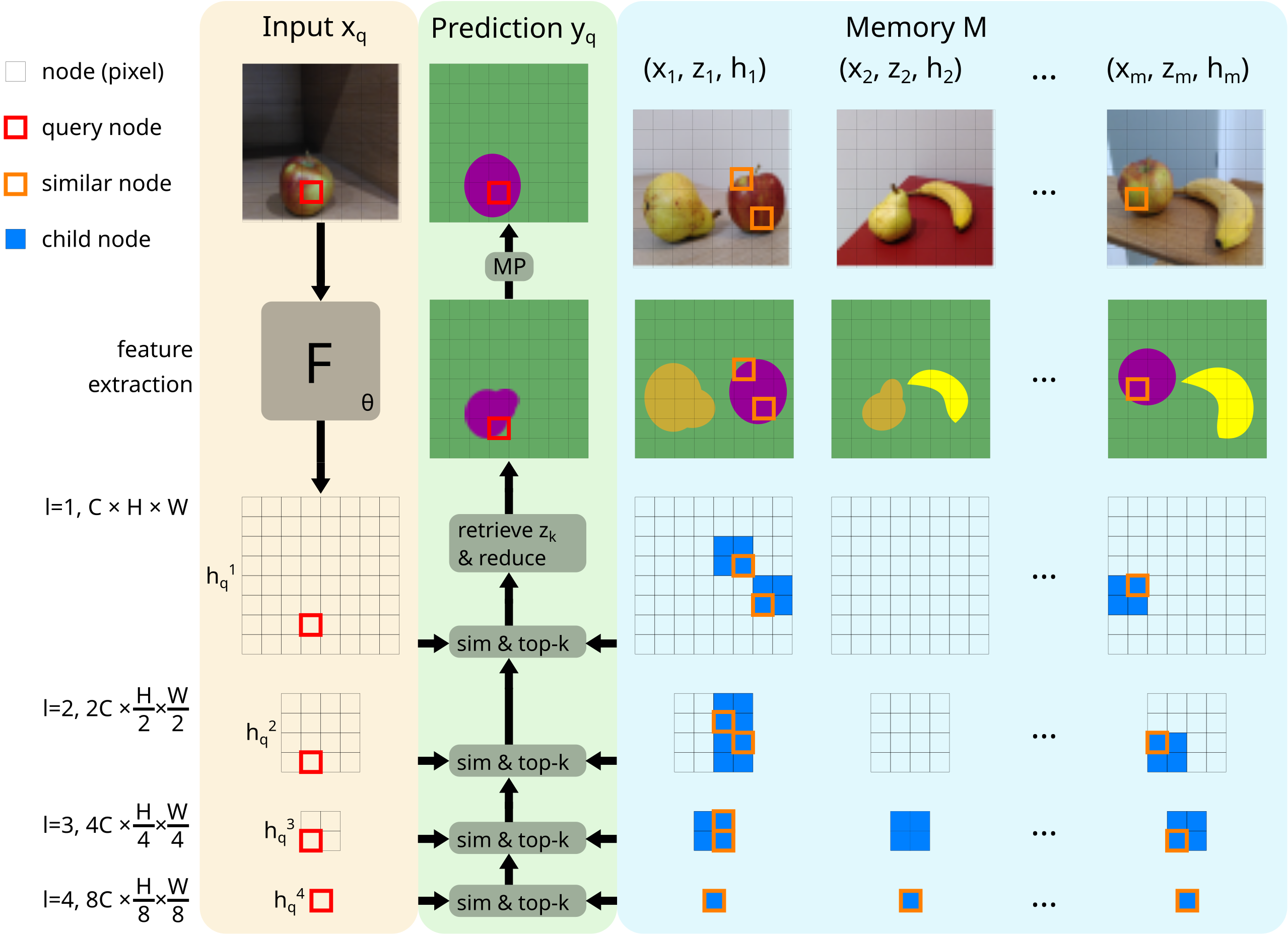}
	\caption{Overview of our parameter-free transduction method for dense prediction tasks. A feature extractor $F$ is applied to a query $x_q$ to obtain hidden representations $(h)_q^l$. A memory $M$ stores labeled samples along with their representations $(h)_j^l$. Our approach then performs a hierarchical search in $M$, where the objective is to find globally and locally similar nodes w.r.t. $x_q$. In each level $l$, we keep the top-$k$ most similar nodes and retrieve their child nodes in level $l-1$. Labels from most similar nodes are retrieved to obtain a raw prediction. Then, message passing (MP) is used to get a final prediction $y_q$.}
	\label{fig:sketch}
\end{figure}

For training task-specific deep learning (DL) models, both data and task are usually well-defined in advance. While this setup holds true in most cases, a major issue emerges when already existing models are required to adapt to new data and new tasks. Satisfying such a flexibility requirement is non-trivial, increasing the difficulty for solving a given problem. Although research in continual learning (CL) resulted in promising approaches, they still suffer from certain restrictions \cite{delange22,wang23} or focus on specific problem-niches \cite{prabhu20}. Moreover, proposed solutions are often unstable \cite{chrysakis20,lee22,delange23} and require domain experts for deployment and maintenance.

For example, a well established CL approach is replay training, which uses a memory module for knowledge retention \cite{shin17,rebuffi17,lopez17,rolnick19}. However, replay training turns out to be challenging in practice due to several reasons. First, models require high plasticity to learn and unlearn continually, which in turn limits their ability to retain knowledge. Forcing models to be robust, i.e., retain knowledge via replay training, is thus somewhat contradicting the plasticity requirement for learning new concepts. Second, all relevant knowledge has to be incorporated into model parameters via training, thereby rendering the available memory capacity irrelevant for inference. Third, the balancing of memory samples and new samples for training is non-trivial in practice and often leads to memory overfitting \cite{lopez17,chaudhry19,verwimp21}. Finally, replay does not allow to unlearn specific examples easily, e.g., in case they are not needed anymore. We argue that simplicity and flexibility are not necessarily mutually exclusive properties of modern DL approaches. In contrast to common, induction-based approaches, where predictions are inferred from general, learned principles, we embrace transduction, which is characterized by reasoning from specific training cases to specific test cases \cite{gammerman98}.

\textbf{Our Work:} We propose PARMESAN (\textbf{par}ameter-free \textbf{me}mory \textbf{s}earch \textbf{a}nd transductio\textbf{n}), an approach that separates computation from memory and extends transduction to dense prediction tasks while at the same time retaining high flexibility. Our method combines several concepts to perform hierarchical correspondence matching with a memory that contains task-specific, labeled data. Rather than requiring to incorporate all knowledge into learnable parameters, the memory in our method extends the capacity of existing models. We further propose a message passing approach to take advantage of correlations within a query sample, as well as a sparsity approach to retain memory efficiency.

PARMESAN is \textit{parameter-free} in the sense that it does not have learnable parameters. Instead, learning can be easily performed via memory consolidation, i.e., adding, removing, or modifying memory samples. Our method allows for fast learning and unlearning of individual examples or parts thereof. Unlike most CL methods, our approach does not require any continual parameter-training by design. It therefore neither suffers from parameter-induced forgetting and memory overfitting, nor does it require complicated and energy-hungry training.

PARMESAN can handle various data dimensionalities such as 1D sequences, 2D images, 3D volumes, as well as spatio-temporal data. We do not impose major restrictions on memory size and allow flexible memory management. Our method can be applied in combination with common DL architectures, thereby enhancing learning speed. This is relevant, e.g., in test-time learning or when dealing with limited computational resources. It is also task-agnostic, i.e., it transfers to different dense prediction tasks as well as multi-tasking.

\noindent
\textbf{Our Contributions:}
\begin{itemize}
	\item A parameter-free transduction method for dense prediction tasks using hierarchical correspondence matching with a memory.
	\item A parameter-free message passing approach to take advantage of intra-query correlations.
	\item A novelty-sparsity approach to achieve high memory diversity while remaining hardware-efficient.
	\item A demonstration of our method's flexibility by using it in combination with common DL archictectures for semantic segmentation and depth estimation using the Cityscapes and JSRT Chest X-ray datasets.
	\item We show that our method can be successfully applied to CL, being on par with well-established baselines in terms of predictive performance and memory efficiency while at the same time learning by 3-4 orders of magnitude faster and having stable knowledge retention.
\end{itemize}

\section{Related work}

Our method is a novel combination of concepts from the following fields:

\textbf{Memory Networks:} Recently, memory networks experience renewed interest \cite{hochreiter97,weston15,graves14,graves16,ramsauer21,paischer23,beck24}. The notion of separating computation and memory \cite{graves16} allows incorporating new information easily while at the same time enabling treatment of memory contents as variables. Memory consolidation \cite{gelbard08,benna16} concerns the strategy for deciding which knowledge to keep in the presence of limited memory resources. A sophisticated memory consolidation method is novelty search \cite{krutsylo22}, where only the most informative examples are kept. We expand on this approach when introducing sparsity to stored samples.

\textbf{Transduction:} Recently, transduction methods experience a revival due to their effectiveness in computer vision \cite{belhasin22}. A well-established transduction method is k-nearest neighbors ($k$-NN) \cite{fix51,jiang23}, which aims to find similar examples w.r.t. common features. Transduction via $k$-NN was demonstrated to work well for image classification \cite{nakata22}, where labeled training samples and their hidden representations are stored in memory. Moreover, $k$-NN in combination with vision transformers \cite{vaswani17,dosovitskiy21} yielded promising results for dense out-of-distribution detection \cite{galesso23}. However, making $k$-NN scalable to dense prediction tasks at high resolutions remains underexplored.

\textbf{Correspondence Matching:} Dense correspondence matching \cite{rocco17,wang19,liu20,lyu20,jiang21,hong21,sun21,edstedt23} is a fundamental problem in computer vision, often acting as a starting point for downstream tasks like optical flow and depth-estimation. In this context, promising methods to improve feature correlation have been proposed \cite{truong20,ashraful23,mariotti24}. Also, feature pyramids were exploited for object detection \cite{girshick15,lin17} and semantic feature matching \cite{ufer17,zhao21}. Recently, foundation models \cite{bomm21} have been applied for feature matching in one-shot semantic segmentation \cite{liu23}. Many existing matching approaches use a single reference sample with known correspondences, but typically can not handle more than one reference sample. Moreover, most methods rely on specific assumptions commonly found in setups with videos or stereo / multi-view images. In contrast, our proposed method allows matching of arbitrary samples and does not require strong assumptions regarding the presence of correspondences.

\textbf{Message Passing (MP):} Graph neural networks \cite{scarselli09,defferrard16,kipf17} introduced the concept of MP to DL, where graph nodes are updated depending on received messages from connected nodes. MP was successfully applied, e.g., to various problems in bioinformatics, material science, and chemistry \cite{zhang21,reiser22}.

\textbf{Continual Learning (CL):} The field of CL \cite{thrun95,ring98,thrun98,parisi19,delange22,wang23,yuan23} studies the ability of an ML model to acquire, update, and accumulate knowledge in an incremental manner. This enables adaptation to data-distribution shifts \cite{quinonero09,tao22} and new tasks. CL is also relevant when frequent re-training is infeasible. Three CL scenarios of increasing difficulty have been proposed \cite{ven19}: task-incremental (TI), domain-incremental (DI), and class-incremental (CI). The definition of the CI scenario is not limited to classification, but is characterized by the ability of a model to infer which task it is presented with.

Prominent CL approaches are based on regularization \cite{kirkpatrick17,li16}, parameter-isolation \cite{mallya18,rusu16}, and replay \cite{shin17,rebuffi17,lopez17,rolnick19,knoblauch20}. In contrast to biological neural networks, replay in ML is used for regularization, where the idea is to learn from new data in tandem with data that is stored in a memory.

In CL, the notion of knowledge retention \cite{grossberg80} plays a major role in learning. Forgetting is affected by aspects such as capacity, plasticity, and the quality as well as diversity of training data. The catastrophic forgetting problem in CL \cite{mccloskey89,french93,mcclelland95,french99,kirkpatrick17} refers to learnable parameters adapting to the most recently seen training data, thereby discarding knowledge from old data. Note that storing sensitive data such as patient information in a memory raises privacy concerns due to the possibility of reconstruction attacks \cite{fredrikson15,sikandar23}. The field of data-free CL \cite{li16} using generated \cite{shin17} or distilled \cite{wangT23,guo23} data has emerged as a viable extension when privacy-preservation is needed.

\section{Method}
In this section we describe our parameter-free transduction method (cf. \cref{fig:sketch}) in more detail. First, we provide an overview of our method and describe the individual components of our model, which consists of a frozen feature extractor, a memory, and a transduction module (\cref{sec:overview}). Given a query input, we perform dense hierarchical correspondence matching with memory contents and take advantage of intra-query correlations via message passing (\cref{sec:corr}). Finally, we improve memory efficiency by introducing sparsity to stored samples (\cref{sec:efficiency}).

\subsection{General Setup of PARMESAN}
\label{sec:overview}

Our proposed method requires a pre-trained feature extractor $F$. Let $F$ be a neural network with learnable parameters $\theta$ consisting of an encoder-like or encoder-decoder-like architecture (e.g., \cite{ronneberger15,woo23,tang21}) that delivers a feature pyramid $(h)^l$ of hidden representations with levels $l=1, ..., n$. We refer to $h^1$ as the hidden representation of $l=1$ (high-res) and $h^{n}$ as the hidden representation of $l=n$ (low-res). Individual ``pixels'' in $(h)^{l}$ with their respective features are called ``nodes''. The total number of nodes for level $l$ is $p^l=\prod_{dim}res_{dim}^l$, where $dim$ refers to the data dimension and $res^l$ to the grid resolutions. Our method is designed to perform CL with frozen parameters $\theta$. $F$ can be pre-trained in a supervised or unsupervised manner, or can be taken from public repositories. Although not strictly required, we recommend pre-training $F$ on a dense prediction task in order to enrich features with dense information.

Let $X$, $H$, and $Z$ be the input-, latent-, and output spaces, respectively. Densely labeled data is referred to as $(x,z)$ with inputs $x\in X$ and labels $z\in Z$. We define a solver $S:X \rightarrow Z$ as the combination of a feature extractor $F:X\rightarrow H$, a memory $M$, and a parameter-free transduction module $G:H \rightarrow Z$. Given a query input $x_q$ and a memory $M$ that stores $m$ labeled samples, $G$ predicts \mbox{$y_q=G(F(x_q; \theta), M)$}. $G$ performs hierarchical correspondence matching between a query feature pyramid $(h)_q^l$ and memory contents, followed by label-retrieval and postprocessing. We use $M$ for both learning and long-term knowledge retention. Knowledge in $M$ can be modified explicitly, allowing for nuanced and sensible management of what to learn, what to retain, and what to forget. Note that this setup shifts the plasticity-robustness trade-off from learnable parameters with parameter-induced forgetting to a memory with memory-induced forgetting. 

In contrast to other approaches, our method is designed to learn fast in the sense that it does not require adapting $\theta$ to perform complex tasks such as CL. Our method only requires memory consolidation, which in our case means saving examples in $M$ and extracting their feature pyramids $(h)_j^l$ in a single forward pass with frozen $\theta$.

We assume grid-based data structures such as images, volumes, and time series. Although we focus on orthogonal grids in this work, our method can be transferred to irregular, adaptive, and isometric grids. We do not impose any restrictions on label-space growth, such as requiring to pre-define the number of output channels, the number of dimensions, or the numerical data type. When taken to extremes, individual tokens of labels can even represent entire data structures such as sets, series, or graphs.

\subsection{Correspondence Matching and Intra-Query Transduction}
\label{sec:corr}
We initialize our model by filling $M$ with $m$ labeled samples $(x,z,(h)^l)_j$, where $j=1,...,m$ and $l=1,...,n$. Leaf nodes in $l=0$ refer to $z_j$. Next, we define connectivity kernels which define children-parent relations across different levels. Kernels remain fixed for every sample and every level (also see \cref{sec:method_appendix}).

\begin{algorithm}[t]
	\scriptsize
	\caption{PARMESAN model inference}
	\label{alg:inference}
	\begin{algorithmic}
		\Require $dim$, $\phi$  \Comment{hyperparameters}
		\Require $x_q, (h)_j^l, z_j\ ; j=1, ..., m\ ; l=1, ..., n$  \Comment{input}
		\State $(h)_q^l \gets F(x_q; \theta)\ ; l=1, ..., n$ \Comment{encode query}
		
		\\ \\ init (in parallel for all $p^n$ query nodes):
		\State $k \gets m$
		\State $id_{k,par} \gets p^n \times [0, 1, 2, ..., k]$  \Comment{parent indices}
		\State $s_{k,par} \gets p^n \times [1., 1., 1., ..., 1.]$  \Comment{parent similarities}
		
		\\ \\ memory search (in parallel for all $p^l$ query nodes):
		\For{\texttt{$l=n, 1$}}
		\State $id_{comp} \gets get\_children(l+1, id_{k,par})$  \Comment{[$p^l \times k \cdot 2^{dim}$]}
		\State $s^l \gets calc\_sim(h_q^l, h_j^l[id_{comp}])$ \Comment{[$p^l\times k \cdot 2^{dim}$]}
		\State $s_{acc}^l \gets s^l \cdot s_{k,par}$  \Comment{update, [$p^l\times k \cdot 2^{dim}$]}
		\If{$l > 1$}
		\State $k \gets \lfloor\phi \cdot k\rfloor$ \Comment{reduce}
		\EndIf
		\State $id_{keep} \gets get\_topk(s_{acc}^l, k)$  \Comment{find top-$k$, [$p^l\times k$]}
		\State $id_{k,par} \gets id_{comp}[id_{keep}]$ \Comment{update, [$p^{l-1}\times k$]}
		\State $s_{k,par} \gets s_{acc}^l[id_{keep}]$ \Comment{update, [$p^{l-1}\times k$]}
		\EndFor
		
		\\ \\label-retrieval and reduction:
		\State $id_{z} \gets get\_children(l=1, id_{k,par})$
		\State $y_{q,raw} \gets sum(softmax(s_{k,par})_k \cdot z_j[id_{z}])_k$
		
		\State return $y_{q,raw}$  \Comment{raw prediction}
	\end{algorithmic}
\end{algorithm}

We formalize the forward pass of our method in \cref{alg:inference}. First, a query $x_q$ is encoded by $F$ to get $(h)_q^l$. Then, $G$ takes $(h)_q^l$ together with stored $(h)_j^l$ and $z_j$ as inputs. The overall objective is to find globally and locally similar patterns in $M$ w.r.t. $x_q$. Traversing levels backwards starting from $l=n$, we compare queries $h_q^l$ with keys $h_j^l$, followed by retrieving the children (within $l-1$) of the top-$k$ most similar nodes of $h_j^l$. We use cosine similarity by default, where the accumulated similarity $s_{acc}^l$ between two nodes at level $l$ is the product of their similarities $s^l$ at level $l$ and the similarities $s_{k,par}$ of their respective parents across all higher levels:

\begin{equation}
	s_{acc}^l=\prod_{u=l}^{n}sim(h_q^u, h_j^u) = \prod_{u=l}^{n}\frac{h_q^u \cdot h_j^u}{\lVert h_q^u\rVert \ \lVert h_j^u\rVert}\ .
\end{equation}

For each of the $p^l$ query nodes, we use $s_{acc}^l$ for comparisons. The top-$k$ most similar nodes in $l$ become parents for $l-1$. Children in $l-1$ are required for processing $l-1$, thereby forming a hierarchy across levels. The reduction rate $\phi \in (0.0,1.0]$ controls $k$, i.e., how many nodes are retained in each level. A variable number of nodes can be retained in different memory samples. Memory search results in a dense correspondence map between a query $x_q$ and, for every query node, their top-$k$ most similar memory nodes $i_{k,par}$. We use these matches to retrieve task-specific labels $z_k$ from $M$. We obtain a raw prediction by weighting retrieved labels with softmax-normalized similarities $s_{k,par}$ of all matches.

Considering the data dimension $dim$ and the grid resolutions $res^l$, exhaustive search would result in a computational complexity of $\mathcal{O}((\prod_{dim} res_{dim}^l)^2)$ for level $l$. In contrast, our method performs local search and has a complexity of only $\mathcal{O}(n_{ch}\cdot(\prod_{dim} res_{dim}^l))$ with $n_{ch}$ being the number of children of nodes in $l+1$. Search is performed independently and in parallel for every query node, making our approach scalable to high resolutions. Moreover, since $h^l$ are strongly correlated across levels, there is a high chance that similarities to query nodes are also strongly correlated across levels. In other words, searching locally pre-filters nodes that are unlikely to be among the top-$k$ most similar nodes in the next level anyway.

Up until this point, no intra-query correlations had been exploited by using geometric or smoothness priors. Strong local correlations are common in grid-based data and should be exploited if possible. We therefore propose a content-aware, parameter-free message passing (MP) approach to account for local correlations within the query and achieve pixel-exact predictions. Re-using our hierarchical search without memory content, top-$\kappa$ nearest neighbor nodes within the query itself are found. Similarity scores between a query node $i$ and its nearest neighbors $j$ are referred to as edges $e_{ij}$. MP is then perfomed to refine raw predictions $y_{q,raw}$. Considering a query node state $y_{q,i}$, we define the equations for the edges $e_{ij}$, the aggregated message $\hat{y}_{q,i}$, and the node update from step $t\rightarrow t+1$ as

\begin{equation}
	e_{ij} = \frac{exp(s_{acc,j}^1)}{\sum_{j=1}^{\kappa} exp(s_{acc,j}^1)} \ ,
\end{equation}

\begin{equation}
	\hat{y}_{q,i} = \sum_{j=1}^{\kappa} e_{ij}\ y_{q,j}^t \ ,
	\label{eq:mp0}
\end{equation}

\begin{equation}
	y_{q,i}^{t+1} = (1 - \lambda) \ y_{q,i}^{t} + \lambda \ \hat{y}_{q,i} \ .
	\label{eq:mp1}
\end{equation}

The hyperparameter $\lambda \in (0.0,1.0]$ controls the overall strength of MP per step. We apply \cref{eq:mp0,eq:mp1} repeatedly for all query nodes in level $l=0$ until convergence \cite{zhou03}, i.e., when $||y_{q,i}^{t+1} - {y}_{q,i}^t|| \rightarrow 0$. The resulting prediction $y_q$ then contains fewer artifacts and is spatially more consistent compared to $y_{q,raw}$. 

\subsection{Memory-Efficiency via Sparsity and Novelty}
\label{sec:efficiency}
\begin{figure}[t]
	\centering
	\includegraphics[width=0.6\linewidth]{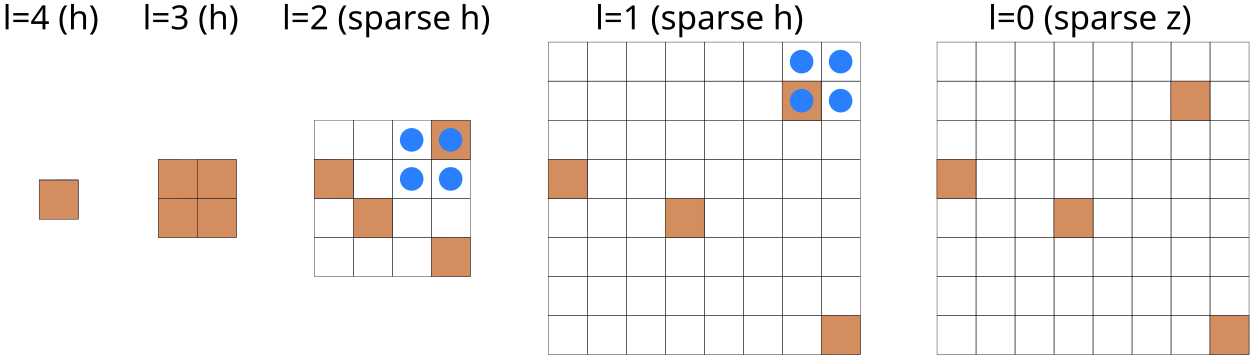}
	\caption{Memory sample feature pyramid with $n_{sp}=2$ sparse levels obtained via iterative, local novelty search (indicated by circles). Filled nodes are kept in $M$.}
	\label{fig:sketch_sparsity}
\end{figure}

In its basic setup, PARMESAN requires considerable hardware memory resources to operate. We thus introduce sparsity, i.e., a reduction of the number of patterns per stored sample, to become exponentially more memory efficient (see \cref{fig:sketch_sparsity}). Sparsity effectively reduces spatial resolutions of stored samples and is applied to both $(h)_j^l$ and $z_j$ after feature extraction of full-resolution inputs $x_j$. Starting from $l=1$, the number of sparse levels $n_{sp}$ defines the overall degree of sparsity. Naive sparsity can be achieved by random subsampling of nodes in levels $l=1,...,n_{sp}$. However, we apply local novelty search to retain high diversity in memory contents. Starting from $l=1$, we keep the most novel node in every $2^{dim}$ patch such that sparsified resolution $res_{sp,dim}^1=res_{dim}^2$. After applying sparsity to $l=2$ such that $res_{sp,dim}^2=res_{dim}^3$, we discard previously kept nodes in $l=1$ such that $res_{sp,dim}^1=res_{sp,dim}^2=res_{dim}^3$. We repeat this procedure for every sparse level and independently for all memory samples.

\section{Experiments and Results}

We demonstrate the capabilities of our method in ablation studies (\cref{sec:ablation_exp}) and CL experiments (\cref{sec:cl_exp}). For ablations, we aim to demonstrate flexibility and efficiency while simplifying and speeding up learning. For CL, we study knowledge retention and overall performance. More details regarding applied models, training and evaluation, and additional results are provided in \cref{sec:datasets_appendix} and \ref{sec:train_and_eval_appendix}.

\textbf{Datasets:} We use the Cityscapes (CITY) dataset \cite{cordts16} for semantic segmentation and monocular depth estimation. Segmentation masks comprise fine and coarse labels. Disparity maps are given for depth estimation. We also use the JSRT Chest X-ray dataset \cite{shiraishi00} for semantic segmentation.

\textbf{Method:} We employ a U-Net variant as our default DL model and feature extractor $F$. Some experiments use a ConvNeXt Tiny \cite{liu22} encoder pre-trained on ADE20K \cite{zhou17}. If $M$ is used, we select memory samples randomly from the training split while keeping the random generation process fixed across all experiments. For PARMESAN, we set $m\smalleq2975$, $\phi\smalleq0.5$, $n_{sp}\smalleq3$ ($m\smalleq199$, $\phi\smalleq0.8$, $n_{sp}\smalleq3$ for JSRT), apply message passing (MP), and take decoder features from $F$ unless stated otherwise. CL is performed by memory consolidation of training samples. We also use test-time augmentation (TTA) to boost performance \cite{jiang21}.

\textbf{Baselines:} We select baselines with the premise to compare to representative and well-established methods in the field. Joint training (JOINT) refers to training on all training data, thus representing the default supervised learning setup and the upper bound for CL \cite{ven19}. Due to catastrophic forgetting, fine-tuning (FT) is regarded as the lower bound for CL \cite{ven19}. We employ classical replay (REPLAY) \cite{shin17,rebuffi17,lopez17,rolnick19}, greedy-sampling dumb learning (GDUMB) \cite{prabhu20}, and Modeling the Background (MiB) \cite{cermelli20} as additional CL baselines. Training via GDUMB is performed from scratch for each individual CL step, thereby exclusively using memory samples. For fair and consistent comparisons, we implement all baselines and adjust method-specific hyperparameters to our data setup.

\textbf{Evaluation:} We use the mean Intersection over Union of classes ($\miou{cl}$) and categories ($\miou{cat}$) as evaluation metrics for semantic segmentation and Root-Mean-Square Error (RMSE) in meters for depth estimation. We report best performances on the CITY validation set and on the JSRT test set, since labels for the CITY test set are not publicly available. For a given hardware setup, the learning speed $\tau_l$ is the average wall-clock time required to learn a training sample. The total learning time for $n$ samples is then $T=n \cdot \tau_l$.

\subsection{Ablation Studies}
\label{sec:ablation_exp}

We perform ablation studies of various components to analyze the impact on learning speed and predictive performance, as well as demonstrating the flexibility of our method with different architectures of $F$ (see \cref{tab:results_ablations_perf}).

\begin{figure}[t]
	\centering
	\includegraphics[width=1.\linewidth]{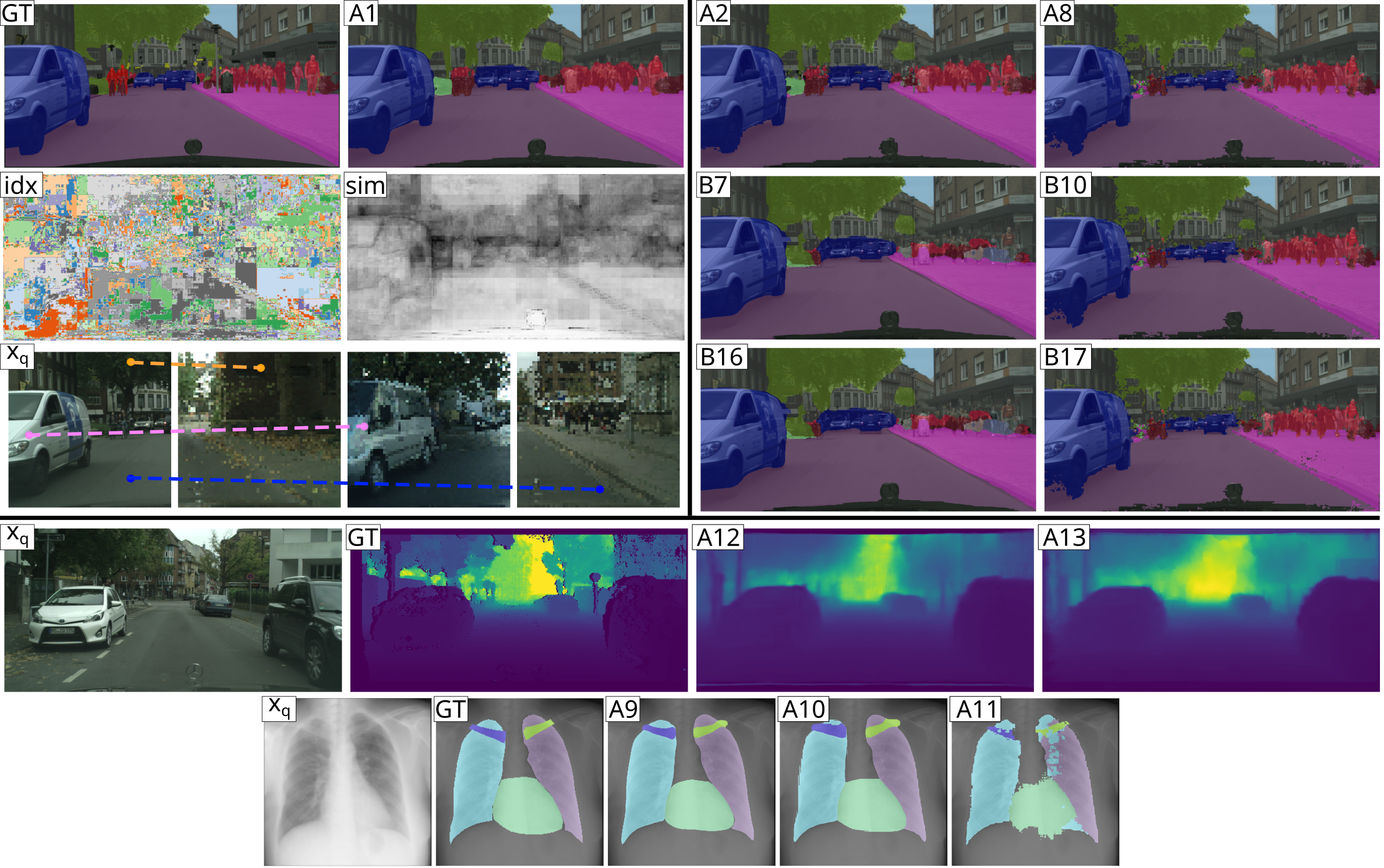}
	\caption{Model predictions and analysis of PARMESAN. \textbf{Left:} We study top-$1$ nearest neighbors from experiment A1. The \textit{idx} panel visualizes indices of retrieved nearest neighbor nodes from $M$. Similar colors indicate labels retrieved from close-by memory locations, most likely from within same samples. The \textit{sim} panel visualizes $s_{acc}^1$, where bright regions refer to high similarities. We also show 3 query pixels and their matches in $M$, indicating both global and local semantic correspondence. \textbf{Right:} Flexible use of PARMESAN in various setups. \textbf{Bottom:} Depth estimation on CITY and semantic segmentation on JSRT.}
	\label{fig:predictions}
\end{figure}

Performance of our method is on-par with JOINT (A0 / A1, A9 / A10, A12 / A13) while greatly enhancing flexibility and learning speed. Learning with PARMESAN can be done in $\tau_l\simeq0.01$ seconds per sample, thereby mitigating the demand for additional computational costs when attached to existing models. Using only the encoder (A4) or even random weights (A5) yields promising results already, indicating that $G$ imposes useful inductive biases. PARMESAN can be successfully used for transfer learning when using a publicly available ConvNeXt pre-trained on ADE20K (A8, A11) or pre-training on coarse labels (A7). MP and TTA (A1 / A2 / A3) considerably improve raw predictions and succeed in reducing small artifacts while retaining sharp corners of predicted segments (see \cref{fig:predictions}).

In general, we observe global and local semantic correspondence of retrieved nearest neighbors from $M$, allowing for future work on various topics related to interpretability and explainability of predictions. Moreover, since every retrieved node comes with its similarity score w.r.t. the query, our method provides pixel-exact indications for uncertainty. As visible in the \textit{sim} panel in \cref{fig:predictions}, similarity scores seem to be correlated with the heterogeneity of regions in the image.

\begin{figure}[!]
	\centering
   \includegraphics[width=0.45\linewidth]{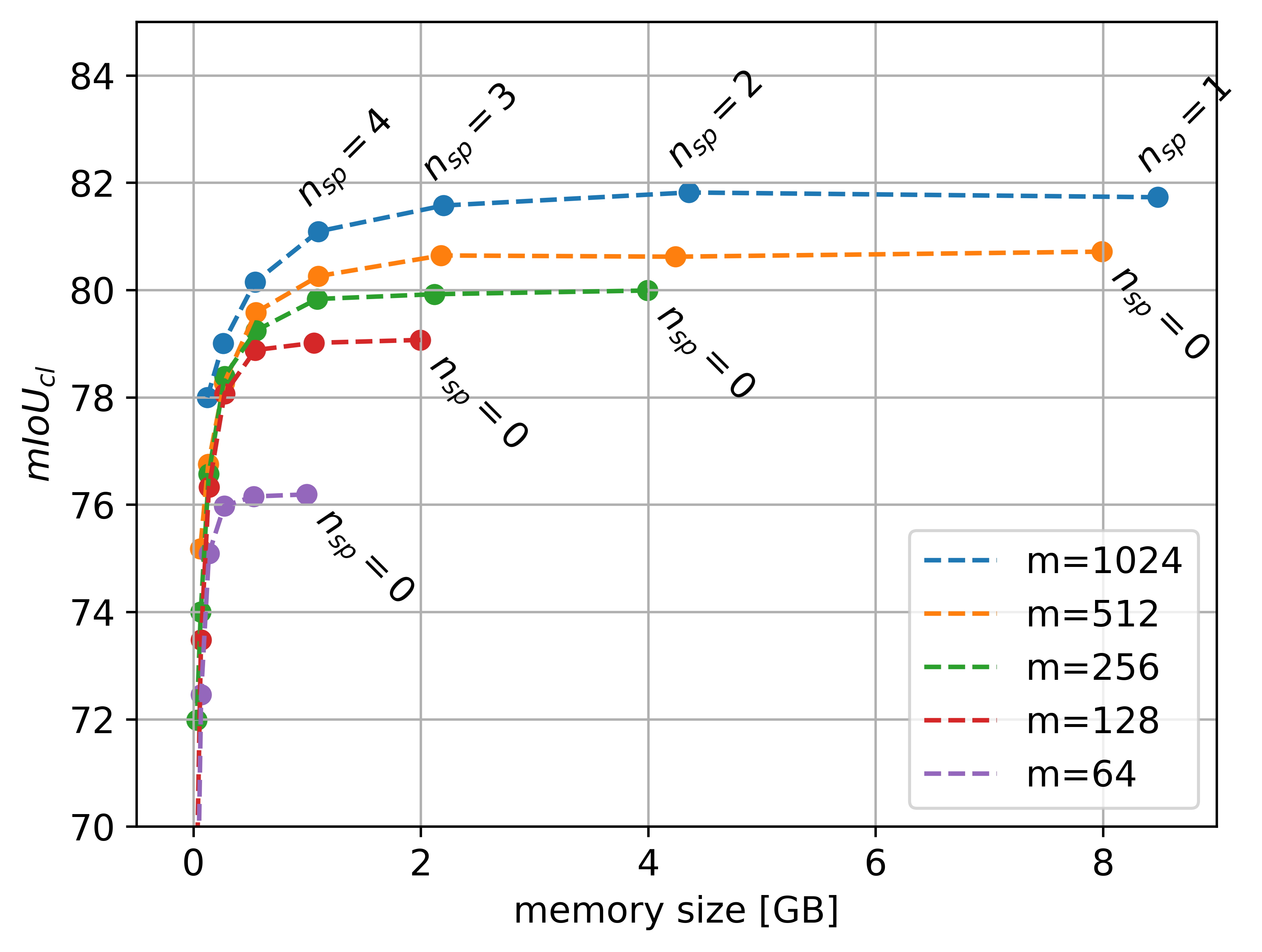}
   \includegraphics[width=0.45\linewidth]{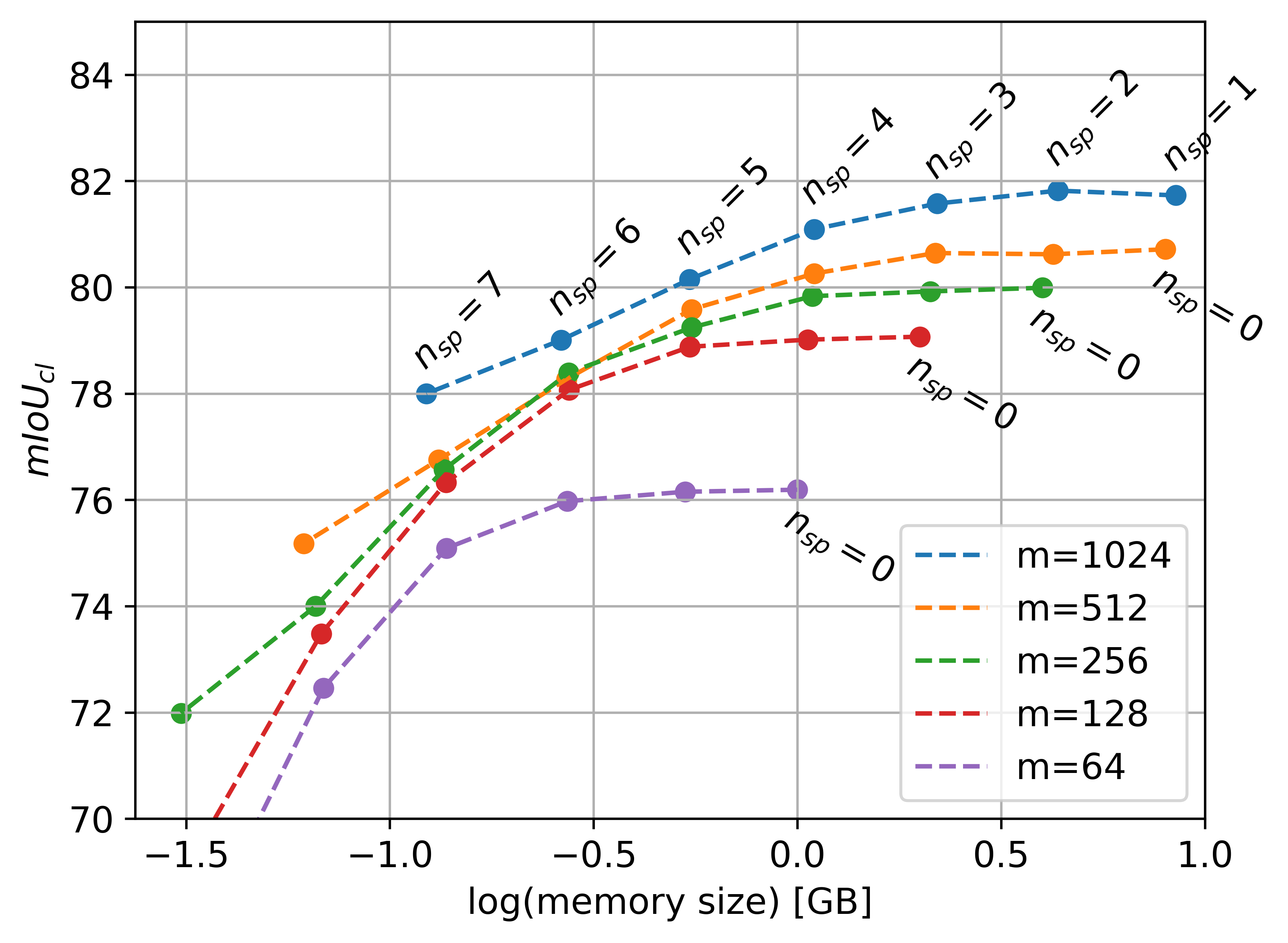}
   \caption{Performance w.r.t. memory size on CITY. We control memory size by introducing sparse levels, leading to exponentially lower memory requirements and stable performance. Relevant contents are retained using local novelty search.}
\label{fig:sparsity_results}
\end{figure}

In \cref{fig:sparsity_results}, we demonstrate that our novelty-sparsity approach leads to exponentially lower memory requirements while achieving robust predictive performance.

\begin{table}[t]
	\scriptsize
	\centering
	\caption{Results for ablation studies, demonstrating the capabilities of PARMESAN. For semantic segmentation, we report performance for classes ($\miou{cl}$) and categories ($\miou{cat}$). ``$\theta$ pre-train'' and ``$\theta$ train'' indicate the usage of a pre-trained feature extractor $F$ and if any additional training of learnable parameters $\theta$ is required. $\tau_l$ is the average wall-clock time for learning a sample.}
	\begin{tabular}{l|l|l|c|c|l|l|l}
		\hline
		ID & Data (\#train-samples) & Method \& Setup & $\theta$ pre-train & $\theta$ train & $\tau_l$ [s] & $\miou{cl}$ & $\miou{cat}$ \\
		\hline
		A0 & CITY, fine (2975) & JOINT & \xmark & \cmark & 3.72 & 82.5 & 87.4 \\
		A1 & CITY, fine (2975) & ours & A0 & \xmark & 0.012 & 82.1 & 87.0 \\
		A2 & CITY, fine (2975) & ours, no MP & A0 & \xmark & 0.012 & 81.4 & 86.4 \\
		A3 & CITY, fine (2975) & ours, no TTA & A0 & \xmark & 0.012 & 81.2 & 86.3 \\
		A4 & CITY, fine (2975) & ours, E only & A0 & \xmark & 0.009 & 75.6 & 81.6 \\
		A5 & CITY, fine (2975) & ours, random $\theta$ & \xmark & \xmark & 0.012 & 59.6 & 66.3 \\
		A6 & CITY, coarse (19998) & JOINT & \xmark & \cmark & 0.43 & 49.0 & 50.3 \\
		A7 & CITY, fine (2975) & ours & A6 & \xmark & 0.012 & 76.9 & 82.4 \\
		A8 & CITY, fine (2975) & ours, ConvNeXt (E) & \cite{liu22} & \xmark & 0.021 & 78.2 & 84.3 \\
		A9 & JSRT (199) & JOINT & \xmark & \cmark & 23.4 & 93.9 & \noneB \\
		A10 & JSRT (199) & ours & A9 & \xmark & 0.009 & 92.6 & \noneB \\
		A11 & JSRT (199) & ours, ConvNeXt (E) & \cite{liu22} & \xmark & 0.011 & 84.7 & \noneB \\
		\hline
		\hline
		ID & Data (\#train-samples) & Method \& Setup & $\theta$ pre-train & $\theta$ train & $\tau_l$ [s] & \multicolumn{2}{c}{RMSE [m]}\\
		\hline
		A12 & CITY, disparity (2975) & JOINT & \xmark & \cmark & 6.71 & \multicolumn{2}{c}{10.5} \\
		A13 & CITY, disparity (2975) & ours & A12 & \xmark & 0.01 & \multicolumn{2}{c}{11.5} \\
		\hline
	\end{tabular}
	\label{tab:results_ablations_perf}
\end{table} 

\subsection{Continual Learning}
\label{sec:cl_exp}

We perform CI-CL experiments, where the goal is to incrementally learn new semantic classes while at the same time automatically inferring the given task. We employ two different CL scenarios: \textit{2-2 (4) cat} and \textit{13-1 (7) cl} \cite{yang23}. At \textit{2-2 (4) cat}, data at every CL step comprises labels for all classes of different categories: $\textit{D}_1$: \textit{void} \& \textit{flat}, $\textit{D}_2$: \textit{construction} \& \textit{object}, $\textit{D}_3$: \textit{nature} \& \textit{sky}, and $\textit{D}_4$: \textit{human} \& \textit{vehicle}. For memory-based methods REPLAY, GDUMB, and PARMESAN, we update samples in $M$ with labels from the current CL step while keeping the stored inputs. REPLAY and GDUMB utilize $M$ for training $\theta$, whereas PARMESAN uses $M$ for transduction. We summarize our results in \cref{tab:results_CL_perf}, \cref{tab:results_CL_retention}, \cref{fig:predictions}, and provide additional information in \cref{sec:CL_appendix}.

FT (B1, B12) and JOINT (B0, B11) behave as expected with strong forgetting and superior performance w.r.t. other methods, respectively. For same memory sizes, PARMESAN is on par with other CL baselines while enhancing flexibility, allowing simple learning and unlearning of individual examples, and boosting learning speed between 3-4 orders of magnitude. We further demonstrate that our method has stable knowledge retention properties (see \cref{tab:results_CL_retention}). This effect can be attributed to our shift towards memory-induced forgetting, where we explicitly state what to learn (new classes), what to retain (old classes), and what to forget (background).

\begin{table}
	\scriptsize
	\centering
	\caption{Results for CL experiments on CITY fine, including size of $M$ and learning times $\tau_l$. Performance is measured on all classes after the final CL step.}
	\begin{tabular}{l|l|l|c|c|c|l|l|l}
		\hline
		ID & scenario & Method \& Setup & $M$ [GB] & $\theta$ pre-train & $\theta$ train & $\tau_l$ [s] & $\miou{cl}$ & $\miou{cat}$ \\
		\hline
		B0 & 2-2 (4) cat & JOINT & \none & A6 & \cmark & 3.64 & 81.6 & 86.6 \\
		B1 & 2-2 (4) cat & FT & \none & A6 & \cmark & 3.44 & 4.1 & 4.3 \\
		B2 & 2-2 (4) cat & REPLAY, $m \smalleq 940$ & 6.43 & A6 & \cmark & 2.79 & 76.8 & 82.6 \\
		B3 & 2-2 (4) cat & GDUMB, $m \smalleq 940$ & 6.43 & A6 & \cmark & 10.7 & 78.9 & 84.5 \\
		B4 & 2-2 (4) cat & MiB & \none & A6 & \cmark & 6.10 & 78.8 & 83.9 \\
		B5 & 2-2 (4) cat & ours, $m \smalleq 256$ & 0.55 & A6 & \xmark & 0.012 & 73.9 & 79.9 \\
		B6 & 2-2 (4) cat & ours, $m \smalleq 940$ & 2.02 & A6 & \xmark & 0.010 & 75.6 & 81.3 \\
		B7 & 2-2 (4) cat & ours & 6.40 & A6 & \xmark & 0.012 & 76.9 & 82.4 \\
		B8 & 2-2 (4) cat & ours, $m \smalleq 256$, ConvNeXt & 0.55 & \cite{liu22} & \xmark & 0.021 & 71.6 & 78.3 \\
		B9 & 2-2 (4) cat & ours, $m \smalleq 940$, ConvNeXt & 2.0 & \cite{liu22} & \xmark & 0.019 & 73.3 & 80.0 \\
		B10 & 2-2 (4) cat & ours, ConvNeXt & 6.35 & \cite{liu22} & \xmark & 0.021 & 78.2 & 84.3 \\
		\hline
		B11 & 13-1 (7) cl & JOINT & \none & A6 & \cmark & 3.97 & 80.4 & 85.2\\
		B12 & 13-1 (7) cl & FT & \none & A6 & \cmark & 1.89 & 0.21 & 0.21\\
		B13 & 13-1 (7) cl & REPLAY, $m \smalleq 940$ & 6.43 & A6 & \cmark & 7.72 & 79.1 & 84.1\\
		B14 & 13-1 (7) cl & GDUMB, $m \smalleq 940$ & 6.43 & A6 & \cmark & 9.36 & 79.1 & 84.1\\
		B15 & 13-1 (7) cl & MiB & \none & A6 & \cmark & 2.44 & 65.2 & 72.5\\
		B16 & 13-1 (7) cl & ours & 6.40 & A6 & \xmark & 0.001 & 76.8 & 82.2\\
		B17 & 13-1 (7) cl & ours, ConvNeXt & 6.35 & \cite{liu22} & \xmark & 0.003 & 78.1 & 84.0\\
		\hline
	\end{tabular}
	\label{tab:results_CL_perf}
\end{table}

\begin{table}[t]
	\scriptsize
	\centering
	\caption{Knowledge retention after 4 CL steps, showing absolute changes $\delta$ between initial and final $\miou{cl}$. Datasets $\textit{D}_i$ comprise semantic classes of different categories according to the respective CL step. Our method has stable knowledge retention and can not suffer from parameter-induced forgetting by design.}
	\begin{tabular}{l|l|l|l|l|l|l|l|l|l|l|l|l|l|l}
		\hline
		& \multicolumn{2}{c|}{REPLAY (B2)} & \multicolumn{2}{c|}{GDUMB (B3)} & \multicolumn{2}{c|}{MiB (B4)} & \multicolumn{2}{c|}{ours (B6)} & \multicolumn{2}{c|}{ours (B7)} & \multicolumn{2}{c|}{ours (B9)} & \multicolumn{2}{c}{ours (B10)}\\
		$\textit{D}_i$ & \multicolumn{1}{c|}{initial} & \multicolumn{1}{c|}{$\delta$} & \multicolumn{1}{c|}{initial} & \multicolumn{1}{c|}{$\delta$} & \multicolumn{1}{c|}{initial} & \multicolumn{1}{c|}{$\delta$} & \multicolumn{1}{c|}{initial} & \multicolumn{1}{c|}{$\delta$} & \multicolumn{1}{c|}{initial} & \multicolumn{1}{c|}{$\delta$} & \multicolumn{1}{c|}{initial} & \multicolumn{1}{c|}{$\delta$} & \multicolumn{1}{c|}{initial} & \multicolumn{1}{c}{$\delta$}\\
		\hline
		$\textit{D}_1$ & 78.70 & -2.80 & 77.89 & -0.71 & 79.73 & -2.60 & 71.21 & -0.79 & 72.05 & -1.03 & 64.39 & -0.95 & 67.99 & -0.45\\
		$\textit{D}_2$ & 49.66 & +3.62 & 55.87 & +0.07 & 57.21 & -3.29 & 49.05 & -0.43 & 50.52 & -0.51 & 50.17 & -0.50 & 54.40 & -0.43\\
		$\textit{D}_3$ & 63.05 & +4.24 & 69.78 & -0.38 & 69.12 & -1.65 & 61.68 & -0.25 & 63.12 & -0.26 & 59.27 & -0.30 & 62.47 & -0.26\\
		$\textit{D}_4$ & 39.60 & \none & 48.37 & \none & 46.67 & \none & 40.33 & \none & 42.03 & \none & 49.27 & \none & 54.68 & \multicolumn{1}{c}{-}\\
		\hline
		\hline
		$avg$ & 57.75 & +1.69 & 62.98 & -0.25 & 63.18 & -2.51 & 55.57 & -0.49 & 56.93 & -0.60 & 55.78 & -0.58 & 59.88 & -0.38\\
		$std$ & 16.95 & 3.90 & 13.32 & 0.39 & 14.35 & 0.82 & 13.62 & 0.27 & 13.29 & 0.39 & 7.31 & 0.33 & 6.57 & 0.10\\
		\hline
	\end{tabular}
	\label{tab:results_CL_retention}
\end{table}

\section{Conclusion}
\textbf{Summary:} We propose PARMESAN, a flexible, parameter-free transduction approach for dense prediction tasks. We also propose a message passing approach to take advantage of intra-query correlations, as well as a novelty-sparsity approach to retain memory efficiency and diversity. Our method is capable of performing CL in an easy and intuitive manner without any continuous training of learnable parameters.
Our approach was demonstrated to learn substantially faster than common baselines while achieving competitive predictive performance and showing stable knowledge retention properties.
\newline\textbf{Limitations:} Depending on hyperparameters like $m$, $\phi$, and $n_{sp}$, PARMESAN can require more computational resources than learned task-heads. Moreover, choosing a suitable feature extractor requires taking a trade-off between a general-purpose model versus a data- and task-specific model.
\newline\textbf{Open Problems:} Possible research opportunities include leveraging self-supervised feature extractors, studying explainability aspects, as well as investigating memory-related topics such as dataset distillation for memory consolidation. Inference speed and memory requirements might be further improved, e.g., by using specialized libraries. Finally, PARMESAN can be beneficial to other fields such as domain adaptation, transfer learning, and few-shot learning, where flexibility and fast learning are often important.

\bibliographystyle{abbrv}
\bibliography{main}

\begin{thebibliography}{100}

\bibitem{beck24}
M.~Beck, K.~P{\"o}ppel, M.~Spanring, A.~Auer, O.~Prudnikova, M.~Kopp,
  G.~Klambauer, J.~Brandstetter, and S.~Hochreiter.
\newblock xlstm: Extended long short-term memory.
\newblock {\em arXiv preprint}, arXiv:2405.04517, 2024.

\bibitem{belhasin22}
O.~Belhasin, G.~Bar-Shalom, and R.~El-Yaniv.
\newblock {TransBoost}: Improving the best {ImageNet} performance using deep
  transduction.
\newblock In A.~H. Oh, A.~Agarwal, D.~Belgrave, and K.~Cho, editors, {\em
  NeurIPS}, volume~35, pages 28363--28373, 2022.

\bibitem{benna16}
M.~K. Benna and S.~Fusi.
\newblock Computational principles of synaptic memory consolidation.
\newblock {\em Nature Neuroscience}, 19(12):1697—1706, 2016.

\bibitem{bomm21}
R.~Bommasani, D.~A. Hudson, E.~Adeli, R.~Altman, S.~Arora, S.~von Arx, M.~S.
  Bernstein, J.~Bohg, A.~Bosselut, E.~Brunskill, et~al.
\newblock On the opportunities and risks of foundation models.
\newblock {\em arXiv preprint}, arXiv:2108.07258, 2021.

\bibitem{cermelli20}
F.~Cermelli, M.~Mancini, S.~R. Bul{\`{o}}, E.~Ricci, and B.~Caputo.
\newblock Modeling the background for incremental learning in semantic
  segmentation.
\newblock In {\em CVPR}, pages 9230--9239, 2020.

\bibitem{chaudhry19}
A.~Chaudhry, M.~Rohrbach, M.~Elhoseiny, T.~Ajanthan, P.~K. Dokania, P.~H.~S.
  Torr, and M.~Ranzato.
\newblock Continual learning with tiny episodic memories.
\newblock In {\em Workshop on Multi-Task and Lifelong Reinforcement Learning},
  2019.

\bibitem{chrysakis20}
A.~Chrysakis and M.-F. Moens.
\newblock Online continual learning from imbalanced data.
\newblock In {\em International Conference on Machine Learning}, 2020.

\bibitem{cordts16}
M.~Cordts, M.~Omran, S.~Ramos, T.~Rehfeld, M.~Enzweiler, R.~Benenson,
  U.~Franke, S.~Roth, and B.~Schiele.
\newblock The cityscapes dataset for semantic urban scene understanding.
\newblock In {\em CVPR}, pages 3213--3223, 2016.

\bibitem{delange22}
M.~De~Lange, R.~Aljundi, M.~Masana, S.~Parisot, X.~Jia, A.~Leonardis,
  G.~Slabaugh, and T.~Tuytelaars.
\newblock A continual learning survey: Defying forgetting in classification
  tasks.
\newblock {\em IEEE TPAMI}, 44(07):3366--3385, 2022.

\bibitem{defferrard16}
M.~Defferrard, X.~Bresson, and P.~Vandergheynst.
\newblock Convolutional neural networks on graphs with fast localized spectral
  filtering.
\newblock In D.~Lee, M.~Sugiyama, U.~Luxburg, I.~Guyon, and R.~Garnett,
  editors, {\em NeurIPS}, volume~29, pages 3837--3845, 2016.

\bibitem{dosovitskiy21}
A.~Dosovitskiy, L.~Beyer, A.~Kolesnikov, D.~Weissenborn, X.~Zhai,
  T.~Unterthiner, M.~Dehghani, M.~Minderer, G.~Heigold, S.~Gelly, J.~Uszkoreit,
  and N.~Houlsby.
\newblock An image is worth 16x16 words: Transformers for image recognition at
  scale.
\newblock In {\em ICLR}, 2021.

\bibitem{edstedt23}
J.~Edstedt, Q.~Sun, G.~B{\"{o}}kman, M.~Wadenb{\"{a}}ck, and M.~Felsberg.
\newblock {RoMa}: Revisiting robust losses for dense feature matching.
\newblock {\em arXiv preprint}, arXiv:2305.15404, 2023.

\bibitem{eigen14}
D.~Eigen, C.~Puhrsch, and R.~Fergus.
\newblock Depth map prediction from a single image using a multi-scale deep
  network.
\newblock In {\em NeurIPS}, volume~27, pages 2366--2374, 2014.

\bibitem{fix51}
E.~Fix and J.~L. Hodges.
\newblock Discriminatory analysis. nonparametric discrimination: Consistency
  properties.
\newblock Technical Report Number 4, USAF School of Aviation Medicine, Randolph
  Field, 1951.

\bibitem{fredrikson15}
M.~Fredrikson, S.~Jha, and T.~Ristenpart.
\newblock Model inversion attacks that exploit confidence information and basic
  countermeasures.
\newblock In {\em 22nd ACM SIGSAC Conference on Computer and Communications
  Security}, CCS '15, pages 1322--1333, 2015.

\bibitem{french93}
R.~M. French.
\newblock Catastrophic interference in connectionist networks: Can it be
  predicted, can it be prevented?
\newblock In {\em NeurIPS}, pages 1176--1177, 1993.

\bibitem{french99}
R.~M. French.
\newblock Catastrophic forgetting in connectionist networks.
\newblock {\em Trends in Cognitive Sciences}, 3(4):128--135, 1999.

\bibitem{galesso23}
S.~Galesso, M.~Argus, and T.~Brox.
\newblock Far away in the deep space: Dense nearest-neighbor-based
  out-of-distribution detection.
\newblock In {\em ICCVW}, 2023.

\bibitem{gammerman98}
A.~Gammerman, V.~Vovk, and V.~Vapnik.
\newblock Learning by transduction.
\newblock In {\em 14th Conference on Uncertainty in Artificial Intelligence},
  UAI'98, pages 148--155, 1998.

\bibitem{gelbard08}
H.~Gelbard-Sagiv, R.~Mukamel, M.~Harel, R.~Malach, and I.~Fried.
\newblock Internally generated reactivation of single neurons in human
  hippocampus during free recall.
\newblock {\em Science}, 322(5898):96--101, 2008.

\bibitem{girshick15}
R.~Girshick, F.~Iandola, T.~Darrell, and J.~Malik.
\newblock Deformable part models are convolutional neural networks.
\newblock In {\em CVPR}, pages 437--446, 2015.

\bibitem{graves14}
A.~Graves, G.~Wayne, and I.~Danihelka.
\newblock Neural turing machines.
\newblock {\em CoRR}, abs/1410.5401, 2014.

\bibitem{graves16}
A.~{Graves}, G.~{Wayne}, M.~{Reynolds}, T.~{Harley}, I.~{Danihelka},
  A.~{Grabska-Barwi{\'n}ska}, S.~G. {Colmenarejo}, E.~{Grefenstette},
  T.~{Ramalho}, J.~{Agapiou}, A.~P. {Badia}, K.~M. {Hermann}, Y.~{Zwols},
  G.~{Ostrovski}, A.~{Cain}, H.~{King}, C.~{Summerfield}, P.~{Blunsom},
  K.~{Kavukcuoglu}, and D.~{Hassabis}.
\newblock Hybrid computing using a neural network with dynamic external memory.
\newblock {\em Nature}, 538(7626):471--476, Oct. 2016.

\bibitem{grossberg80}
S.~Grossberg.
\newblock How does a brain build a cognitive code?
\newblock {\em Psychological Review}, 87:1--51, 1980.

\bibitem{guo23}
Z.~Guo, K.~Wang, G.~Cazenavette, H.~Li, K.~Zhang, and Y.~You.
\newblock Towards lossless dataset distillation via difficulty-aligned
  trajectory matching.
\newblock {\em CoRR}, abs/2310.05773, 2023.

\bibitem{he16}
K.~He, X.~Zhang, S.~Ren, and J.~Sun.
\newblock Deep residual learning for image recognition.
\newblock In {\em CVPR}, pages 770--778, 2016.

\bibitem{hochreiter97}
S.~Hochreiter and J.~Schmidhuber.
\newblock Long short-term memory.
\newblock {\em Neural Computation}, 9:1735–1780, 1997.

\bibitem{hong21}
S.~Hong and S.~Kim.
\newblock Deep matching prior: Test-time optimization for dense correspondence.
\newblock In {\em ICCV}, pages 9907--9917, October 2021.

\bibitem{ashraful23}
A.~Islam, B.~Lundell, H.~Sawhney, S.~N. Sinha, P.~Morales, and R.~J. Radke.
\newblock Self-supervised learning with local contrastive loss for detection
  and semantic segmentation.
\newblock In {\em IEEE/CVF Winter Conference on Applications of Computer Vision
  (WACV)}, pages 5613--5622, 2023.

\bibitem{jaccard12}
P.~Jaccard.
\newblock The distribution of the flora in the alpine zone.
\newblock {\em New Phytologist}, 11(2):37--50, 1912.

\bibitem{jiang21}
W.~Jiang, E.~Trulls, J.~Hosang, A.~Tagliasacchi, and K.~M. Yi.
\newblock {COTR: Correspondence Transformer for Matching Across Images}.
\newblock In {\em ICCV}, pages 6207--6217, 2021.

\bibitem{jiang23}
Z.~Jiang, M.~Yang, M.~Tsirlin, R.~Tang, Y.~Dai, and J.~Lin.
\newblock {``}{Low}-resource{''} text classification: A parameter-free
  classification method with compressors.
\newblock In {\em Findings of the Association for Computational Linguistics:
  ACL 2023}, pages 6810--6828, 2023.

\bibitem{kelley60}
H.~J. Kelley.
\newblock Gradient theory of optimal flight paths.
\newblock {\em Ars Journal}, 30(10):947--954, 1960.

\bibitem{kingma14}
D.~P. Kingma and J.~Ba.
\newblock Adam: A method for stochastic optimization.
\newblock In Y.~Bengio and Y.~LeCun, editors, {\em ICLR}, 2015.

\bibitem{kipf17}
T.~N. Kipf and M.~Welling.
\newblock Semi-supervised classification with graph convolutional networks.
\newblock In {\em ICLR}, 2017.

\bibitem{kirkpatrick17}
J.~Kirkpatrick, R.~Pascanu, N.~Rabinowitz, J.~Veness, G.~Desjardins, A.~A.
  Rusu, K.~Milan, J.~Quan, T.~Ramalho, A.~Grabska-Barwinska, D.~Hassabis,
  C.~Clopath, D.~Kumaran, and R.~Hadsell.
\newblock Overcoming catastrophic forgetting in neural networks.
\newblock {\em Proceedings of the National Academy of Sciences},
  114(13):3521--3526, 2017.

\bibitem{knoblauch20}
J.~Knoblauch, H.~Husain, and T.~Diethe.
\newblock Optimal continual learning has perfect memory and is {NP-HARD}.
\newblock In {\em ICML}, ICML'20, pages 5327--5337, 2020.

\bibitem{krutsylo22}
A.~Krutsylo and P.~Morawiecki.
\newblock Diverse memory for experience replay in continual learning.
\newblock In {\em 30th European Symposium on Artificial Neural Networks,
  Computational Intelligence and Machine Learning (ESANN)}, pages 91--96, 2022.

\bibitem{delange23}
M.~D. Lange, G.~M. van~de Ven, and T.~Tuytelaars.
\newblock Continual evaluation for lifelong learning: Identifying the stability
  gap.
\newblock In {\em The Eleventh International Conference on Learning
  Representations}, 2023.

\bibitem{lee22}
K.-Y. Lee, Y.~Zhong, and Y.-X. Wang.
\newblock Do pre-trained models benefit equally in continual learning?
\newblock {\em 2023 IEEE/CVF Winter Conference on Applications of Computer
  Vision (WACV)}, pages 6474--6482, 2022.

\bibitem{li16}
Z.~Li and D.~Hoiem.
\newblock Learning without forgetting.
\newblock In B.~Leibe, J.~Matas, N.~Sebe, and M.~Welling, editors, {\em ECCV},
  pages 614--629, 2016.

\bibitem{lin17}
T.-Y. Lin, P.~Doll{\'a}r, R.~Girshick, K.~He, B.~Hariharan, and S.~Belongie.
\newblock Feature pyramid networks for object detection.
\newblock In {\em CVPR}, pages 2117--2125, 2017.

\bibitem{liu20}
Y.~Liu, L.~Zhu, M.~Yamada, and Y.~Yang.
\newblock Semantic correspondence as an optimal transport problem.
\newblock In {\em CVPR}, pages 4463--4472, 2020.

\bibitem{liu23}
Y.~Liu, M.~Zhu, H.~Li, H.~Chen, X.~Wang, and C.~Shen.
\newblock Matcher: Segment anything with one shot using all-purpose feature
  matching.
\newblock {\em arXiv preprint}, 2023.

\bibitem{liu22}
Z.~Liu, H.~Mao, C.-Y. Wu, C.~Feichtenhofer, T.~Darrell, and S.~Xie.
\newblock A {ConvNet} for the 2020s.
\newblock In {\em CVPR}, pages 11976--11983, 2022.

\bibitem{lopez17}
D.~Lopez-Paz and M.~Ranzato.
\newblock Gradient episodic memory for continual learning.
\newblock In {\em NeurIPS}, pages 6470--6479, 2017.

\bibitem{loshchilov18}
I.~Loshchilov and F.~Hutter.
\newblock Decoupled weight decay regularization.
\newblock In {\em ICLR}, 2019.

\bibitem{lyu20}
W.~Lyu, L.~Chen, Z.~Zhou, and W.~Wu.
\newblock Deep semantic feature matching using confidential correspondence
  consistency.
\newblock {\em IEEE Access}, 8:12802--12814, 2020.

\bibitem{mallya18}
A.~Mallya and S.~Lazebnik.
\newblock {PackNet}: Adding multiple tasks to a single network by iterative
  pruning.
\newblock In {\em CVPR}, pages 7765--7773, 2018.

\bibitem{mariotti24}
O.~Mariotti, O.~Mac~Aodha, and H.~Bilen.
\newblock Improving semantic correspondence with viewpoint-guided spherical
  maps.
\newblock In {\em CVPR}, pages 19521--19530, June 2024.

\bibitem{mcclelland95}
J.~McClelland, B.~Mcnaughton, and R.~O’Reilly.
\newblock Why there are complementary learning systems in the hippocampus and
  neocortex: Insights from the successes and failures of connectionist models
  of learning and memory.
\newblock {\em Psychological Review}, 102:419--57, 1995.

\bibitem{mccloskey89}
M.~McCloskey and N.~J. Cohen.
\newblock Catastrophic interference in connectionist networks: The sequential
  learning problem.
\newblock In G.~H. Bower, editor, {\em {Advances in Research and Theory}},
  volume~24 of {\em Psychology of Learning and Motivation}, pages 109--165.
  1989.

\bibitem{morgan89}
N.~Morgan and H.~Bourlard.
\newblock Generalization and parameter estimation in feedforward nets: Some
  experiments.
\newblock In D.~Touretzky, editor, {\em NeurIPS}, volume~2, pages 630--637.
  Morgan-Kaufmann, 1989.

\bibitem{nakata22}
K.~Nakata, Y.~Ng, D.~Miyashita, A.~Maki, Y.-C. Lin, and J.~Deguchi.
\newblock Revisiting a {kNN}-based image classification system with
  high-capacity storage.
\newblock In S.~Avidan, G.~Brostow, M.~Ciss{\'e}, G.~M. Farinella, and
  T.~Hassner, editors, {\em ECCV}, pages 457--474, 2022.

\bibitem{paischer23}
F.~Paischer, T.~Adler, M.~Hofmarcher, and S.~Hochreiter.
\newblock Semantic helm: A human-readable memory for reinforcement learning,
  2023.

\bibitem{parisi19}
G.~I. Parisi, R.~Kemker, J.~L. Part, C.~Kanan, and S.~Wermter.
\newblock Continual lifelong learning with neural networks: A review.
\newblock {\em Neural Networks}, 113:54--71, 2019.

\bibitem{pascanu13}
R.~Pascanu, T.~Mikolov, and Y.~Bengio.
\newblock On the difficulty of training recurrent neural networks.
\newblock In {\em ICML}, ICML'13, pages 1310--1318, 2013.

\bibitem{paszke19}
A.~Paszke, S.~Gross, F.~Massa, A.~Lerer, J.~Bradbury, G.~Chanan, T.~Killeen,
  Z.~Lin, N.~Gimelshein, L.~Antiga, A.~Desmaison, A.~Kopf, E.~Yang, Z.~DeVito,
  M.~Raison, A.~Tejani, S.~Chilamkurthy, B.~Steiner, L.~Fang, J.~Bai, and
  S.~Chintala.
\newblock {PyTorch}: An imperative style, high-performance deep learning
  library.
\newblock In H.~Wallach, H.~Larochelle, A.~Beygelzimer, F.~d\textquotesingle
  Alch\'{e}-Buc, E.~Fox, and R.~Garnett, editors, {\em NeurIPS}, pages
  8024--8035. 2019.

\bibitem{polyak90}
B.~Polyak.
\newblock New stochastic approximation type procedures.
\newblock {\em Avtomatica i Telemekhanika}, 7:98--107, 1990.

\bibitem{prabhu20}
A.~Prabhu, P.~Torr, and P.~Dokania.
\newblock Gdumb: A simple approach that questions our progress in continual
  learning.
\newblock In {\em ECCV}, pages 524--540, 2020.

\bibitem{quinonero09}
J.~Quinonero-Candela, M.~Sugiyama, A.~Schwaighofer, and N.~D. Lawrence,
  editors.
\newblock {\em Dataset Shift in Machine Learning}.
\newblock MIT Press, 2009.

\bibitem{ramsauer21}
H.~Ramsauer, B.~Sch{\"a}fl, J.~Lehner, P.~Seidl, M.~Widrich, L.~Gruber,
  M.~Holzleitner, T.~Adler, D.~Kreil, M.~K. Kopp, G.~Klambauer,
  J.~Brandstetter, and S.~Hochreiter.
\newblock Hopfield networks is all you need.
\newblock In {\em ICLR}, 2021.

\bibitem{rebuffi17}
S.~Rebuffi, A.~Kolesnikov, G.~Sperl, and C.~H. Lampert.
\newblock {iCaRL}: Incremental classifier and representation learning.
\newblock In {\em CVPR}, pages 5533--5542, 2017.

\bibitem{reiser22}
P.~Reiser, M.~Neubert, A.~Eberhard, L.~Torresi, C.~Zhou, C.~Shao, H.~Metni,
  C.~{van Hoesel}, H.~Schopmans, T.~Sommer, and P.~Friederich.
\newblock Graph neural networks for materials science and chemistry.
\newblock {\em Communications Materials}, 3:93, 2022.

\bibitem{ring98}
M.~B. Ring.
\newblock Child: A first step towards continual learning.
\newblock In S.~Thrun and L.~Pratt, editors, {\em Learning to Learn}, pages
  261--292. 1998.

\bibitem{robbins51}
H.~Robbins and S.~Monro.
\newblock {A Stochastic Approximation Method}.
\newblock {\em The Annals of Mathematical Statistics}, 22(3):400 -- 407, 1951.

\bibitem{rocco17}
I.~Rocco, R.~Arandjelovic, and J.~Sivic.
\newblock Convolutional neural network architecture for geometric matching.
\newblock In {\em CVPR}, pages 39--48, 2017.

\bibitem{rolnick19}
D.~Rolnick, A.~Ahuja, J.~Schwarz, T.~Lillicrap, and G.~Wayne.
\newblock Experience replay for continual learning.
\newblock In H.~Wallach, H.~Larochelle, A.~Beygelzimer, F.~d\textquotesingle
  Alch\'{e}-Buc, E.~Fox, and R.~Garnett, editors, {\em NeurIPS}, volume~32,
  pages 348--358, 2019.

\bibitem{ronneberger15}
O.~Ronneberger, P.~Fischer, and T.~Brox.
\newblock {U-Net}: Convolutional networks for biomedical image segmentation.
\newblock In N.~Navab, J.~Hornegger, W.~M. Wells, and A.~F. Frangi, editors,
  {\em Medical Image Computing and Computer-Assisted Intervention (MICCAI)},
  volume 9351, pages 234--241, 2015.

\bibitem{rumelhart86}
D.~Rumelhart, G.~E. Hinton, and R.~Williams.
\newblock Learning representations by back-propagating errors.
\newblock {\em Nature}, 323:533–536, 1986.

\bibitem{ruppert88}
D.~Ruppert.
\newblock Efficient estimations from a slowly convergent robbins-monro process.
\newblock Technical Report No. 781, Cornell University Operations Research and
  Industrial Engineering, 1988.

\bibitem{rusu16}
A.~A. Rusu, N.~C. Rabinowitz, G.~Desjardins, H.~Soyer, J.~Kirkpatrick,
  K.~Kavukcuoglu, R.~Pascanu, and R.~Hadsell.
\newblock Progressive neural networks.
\newblock {\em CoRR}, abs/1606.04671, 2016.

\bibitem{woo23}
R.~H. X. C. Z. L. I. S.~K. Sanghyun~Woo, Shoubhik~Debnath and S.~Xie.
\newblock Convnext v2: Co-designing and scaling convnets with masked
  autoencoders.
\newblock {\em arXiv preprint}, arXiv:2301.00808, 2023.

\bibitem{scarselli09}
F.~Scarselli, M.~Gori, A.~C. Tsoi, M.~Hagenbuchner, and G.~Monfardini.
\newblock The graph neural network model.
\newblock {\em IEEE Transactions on Neural Networks}, 20(1):61--80, 2009.

\bibitem{shin17}
H.~Shin, J.~K. Lee, J.~Kim, and J.~Kim.
\newblock Continual learning with deep generative replay.
\newblock In {\em NeurIPS}, pages 2994–--3003, 2017.

\bibitem{shiraishi00}
J.~Shiraishi, S.~Katsuragawa, J.~Ikezoe, T.~Matsumoto, T.~Kobayashi, K.-I.
  Komatsu, M.~Matsui, H.~Fujita, Y.~Kodera, and K.~Doi.
\newblock Development of a digital image database for chest radiographs with
  and without a lung nodule: Receiver operating characteristic analysis of
  radiologists' detection of pulmonary nodules.
\newblock {\em American Journal of Roentgenology}, 174:71--74, 2000.

\bibitem{sikandar23}
H.~S. Sikandar, H.~Waheed, S.~Tahir, S.~U.~R. Malik, and W.~Rafique.
\newblock A detailed survey on federated learning attacks and defenses.
\newblock {\em Electronics}, 12(2):260, 2023.

\bibitem{sun21}
J.~Sun, Z.~Shen, Y.~Wang, H.~Bao, and X.~Zhou.
\newblock {LoFTR}: Detector-free local feature matching with transformers.
\newblock {\em CVPR}, pages 8922--8931, 2021.

\bibitem{tao22}
T.~Sun, M.~Segu, J.~Postels, Y.~Wang, L.~Van~Gool, B.~Schiele, F.~Tombari, and
  F.~Yu.
\newblock {SHIFT}: A synthetic driving dataset for continuous multi-task domain
  adaptation.
\newblock In {\em CVPR}, pages 21371--21382, June 2022.

\bibitem{tang21}
Y.~Tang, D.~Yang, W.~Li, H.~R. Roth, B.~A. Landman, D.~Xu, V.~Nath, and
  A.~Hatamizadeh.
\newblock Self-supervised pre-training of swin transformers for 3d medical
  image analysis.
\newblock {\em 2022 IEEE/CVF Conference on Computer Vision and Pattern
  Recognition (CVPR)}, pages 20698--20708, 2021.

\bibitem{tanimoto58}
T.~Tanimoto.
\newblock {\em An Elementary Mathematical Theory of Classification and
  Prediction}.
\newblock International Business Machines Corporation, 1958.

\bibitem{tarvainen17}
A.~Tarvainen and H.~Valpola.
\newblock Mean teachers are better role models: Weight-averaged consistency
  targets improve semi-supervised deep learning results.
\newblock In {\em NeurIPS}, pages 1195--1204, 2017.

\bibitem{thrun98}
S.~Thrun.
\newblock Lifelong learning algorithms.
\newblock In S.~Thrun and L.~Pratt, editors, {\em Learning to Learn}, pages
  181--209. 1998.

\bibitem{thrun95}
S.~Thrun and T.~M. Mitchell.
\newblock Lifelong robot learning.
\newblock In L.~Steels, editor, {\em The Biology and Technology of Intelligent
  Autonomous Agents}, pages 165--196, 1995.

\bibitem{truong20}
P.~Truong, M.~Danelljan, L.~V. Gool, and R.~Timofte.
\newblock {GOCor}: Bringing globally optimized correspondence volumes into your
  neural network.
\newblock In {\em NeurIPS}, pages 14278--14290, 2020.

\bibitem{ufer17}
N.~Ufer and B.~Ommer.
\newblock Deep semantic feature matching.
\newblock In {\em CVPR}, pages 5929--5938, 2017.

\bibitem{ulyanov16}
D.~Ulyanov, A.~Vedaldi, and V.~S. Lempitsky.
\newblock Instance normalization: The missing ingredient for fast stylization.
\newblock {\em CoRR}, abs/1607.08022, 2016.

\bibitem{ven19}
G.~M. van~de Ven and A.~S. Tolias.
\newblock Three scenarios for continual learning.
\newblock {\em CoRR}, abs/1904.07734, 2019.

\bibitem{vaswani17}
A.~Vaswani, N.~Shazeer, N.~Parmar, J.~Uszkoreit, L.~Jones, A.~N. Gomez,
  {\L}.~Kaiser, and I.~Polosukhin.
\newblock Attention is all you need.
\newblock In I.~Guyon, U.~V. Luxburg, S.~Bengio, H.~Wallach, R.~Fergus,
  S.~Vishwanathan, and R.~Garnett, editors, {\em NeurIPS}, volume~30, pages
  5998--6008, 2017.

\bibitem{verwimp21}
E.~Verwimp, M.~De~Lange, and T.~Tuytelaars.
\newblock Rehearsal revealed: The limits and merits of revisiting samples in
  continual learning.
\newblock In {\em ICCV}, pages 9385--9394, 2021.

\bibitem{wang23}
L.~Wang, X.~Zhang, H.~Su, and J.~Zhu.
\newblock A comprehensive survey of continual learning: Theory, method and
  application.
\newblock {\em IEEE Transactions on Pattern Analysis and Machine Intelligence},
  pages 1--20, 2024.

\bibitem{wangT23}
T.~Wang, J.~Zhu, A.~Torralba, and A.~A. Efros.
\newblock Dataset distillation.
\newblock {\em CoRR}, abs/1811.10959, 2018.

\bibitem{wang19}
X.~Wang.
\newblock {\em Learning and Reasoning with Visual Correspondence in Time}.
\newblock PhD thesis, Carnegie Mellon University, Pittsburgh, PA, September
  2019.

\bibitem{weston15}
J.~Weston, S.~Chopra, and A.~Bordes.
\newblock Memory networks.
\newblock In {\em ICLR}, 2015.

\bibitem{yang23}
G.~Yang, E.~Fini, D.~Xu, P.~Rota, M.~Ding, M.~Nabi, X.~Alameda-Pineda, and
  E.~Ricci.
\newblock Uncertainty-aware contrastive distillation for incremental semantic
  segmentation.
\newblock {\em IEEE Transactions on Pattern Analysis and Machine Intelligence},
  45(2):2567--2581, 2023.

\bibitem{yuan23}
B.~Yuan and D.~Zhao.
\newblock A survey on continual semantic segmentation: Theory, challenge,
  method and application, 2023.

\bibitem{zhang21}
X.-M. Zhang, L.~Liang, L.~Liu, and M.-J. Tang.
\newblock Graph neural networks and their current applications in
  bioinformatics.
\newblock {\em Frontiers in Genetics}, 12:690049, 2021.

\bibitem{zhao21}
D.~Zhao, Z.~Song, Z.~Ji, G.~Zhao, W.~Ge, and Y.~Yu.
\newblock Multi-scale matching networks for semantic correspondence.
\newblock In {\em ICCV}, pages 3354--3364, 2021.

\bibitem{zhou17}
B.~Zhou, H.~Zhao, X.~Puig, S.~Fidler, A.~Barriuso, and A.~Torralba.
\newblock Scene parsing through {ADE20K} dataset.
\newblock In {\em CVPR}, pages 5122--5130, July 2017.

\bibitem{zhou03}
D.~Zhou, O.~Bousquet, T.~N. Lal, J.~Weston, and B.~Sch\"{o}lkopf.
\newblock Learning with local and global consistency.
\newblock In {\em NeurIPS}, page 321–328, 2003.

\end{thebibliography}

\newpage
\appendix
\setcounter{page}{1}
\title{Supplementary Material}
\subtitle{PARMESAN: Parameter-Free Memory Search and Transduction for Dense Prediction Tasks}

\author{Philip Matthias Winter*, Maria Wimmer*, David Major, Dimitrios Lenis, Astrid Berg, Theresa Neubauer, Gaia Romana De Paolis, Johannes Novotny, Sophia Ulonska, Katja Bühler}
\authorrunning{P.M.~Winter, M.~Wimmer et al.}

\institute{VRVis GmbH, Donau-City-Straße 11, 1220 Vienna\\
\email{office@vrvis.at}\\
\url{https://www.vrvis.at}\\
*These authors contributed equally to this work.}

\maketitle

\section{Method}
\label{sec:method_appendix}
\textbf{Initialization:} We now describe the initialization of $G$ in more detail. First, we need to initialize hyperparameters such as the number of data dimensions $dim$, the number of memory samples $m$ and the resolutions $(res)^l$ for all levels. We then define connectivity kernels between nodes of different levels, which remain fixed for all samples. Each node in level $l$ has a parent node in level $l+1$. For the highest level $l=n$, all nodes have a single root node as their parent. Each node in level $l$ typically has several child nodes in level $l-1$, e.g., $n_{ch}=2^{dim}$. Leaf nodes in level $l=0$ refer to memory labels $z_j$. We assume that resolutions of $h^1$, $x$, and $z$ match, implying that each node in level $l=1$ has exactly one child node in $l=0$. We initialize parent indices $i_{k,par}$ such that hierarchical search has access to the full memory content. We initialize parent similarity scores $s_{k,par}$ with a value of 1.0 for all nodes.

Some neural architectures may not provide desired feature pyramids in the sense that resolutions are halved at every level or that bottlenecks have small resolutions. In these cases, connectivity kernels can either be adjusted accordingly, or missing representations can be inserted by interpolation or extrapolation of neighboring representations. 

\textbf{Technical Aspects:} In order to avoid increasingly exhaustive search due to a rapidly growing number of nodes, we set $\phi$ such that hardware memory requirements remain quasi-constant across levels. For example, $\phi=0.5$ for 2D data in combination with doubling the channel dimension in each level yields quasi-constant hardware memory usage. The same is true for 3D data and $\phi=0.25$. This is especially relevant in case of high-dimensional data and high resolutions, where available hardware often imposes hard limits on the choice of hyperparameters. Naturally, smaller $m$ allow for larger $\phi$. For depth estimation with PARMESAN, we ignore invalid labels in $M$ by manually setting their respective similarity scores $s_{k,par}=-100.0$ before applying the \textit{softmax} operation.

We implemented all methods and experiments using the PyTorch framework~\cite{paszke19}. We apply the exact same pre and postprocessing to retain comparability. We conduct all experiments on a single NVIDIA GeForce RTX 3090 (24 GB) graphics card. PARMESAN has inference times in the order of 1 second per image for $m=256$, $\phi=0.5$, and an input resolution of 512$\times$1024 pixels. By default, we store memory content on RAM and transfer representations to GPU when needed for hierarchical search. In case of small GPU RAM, the hierarchical search can be chunked into smaller pieces via windowing and performed sequentially without compromising performance. For MP, all operations can be done on GPU.

\section{Experimental Setup and Additional Experiments}
\label{sec:experiments_appendix}

In this section, we provide more details regarding our overall experimtal setup and perform some additional experiments.

\subsection{Datasets}
\label{sec:datasets_appendix}
\begin{figure}[t]
\centering
   \includegraphics[width=1.\linewidth]{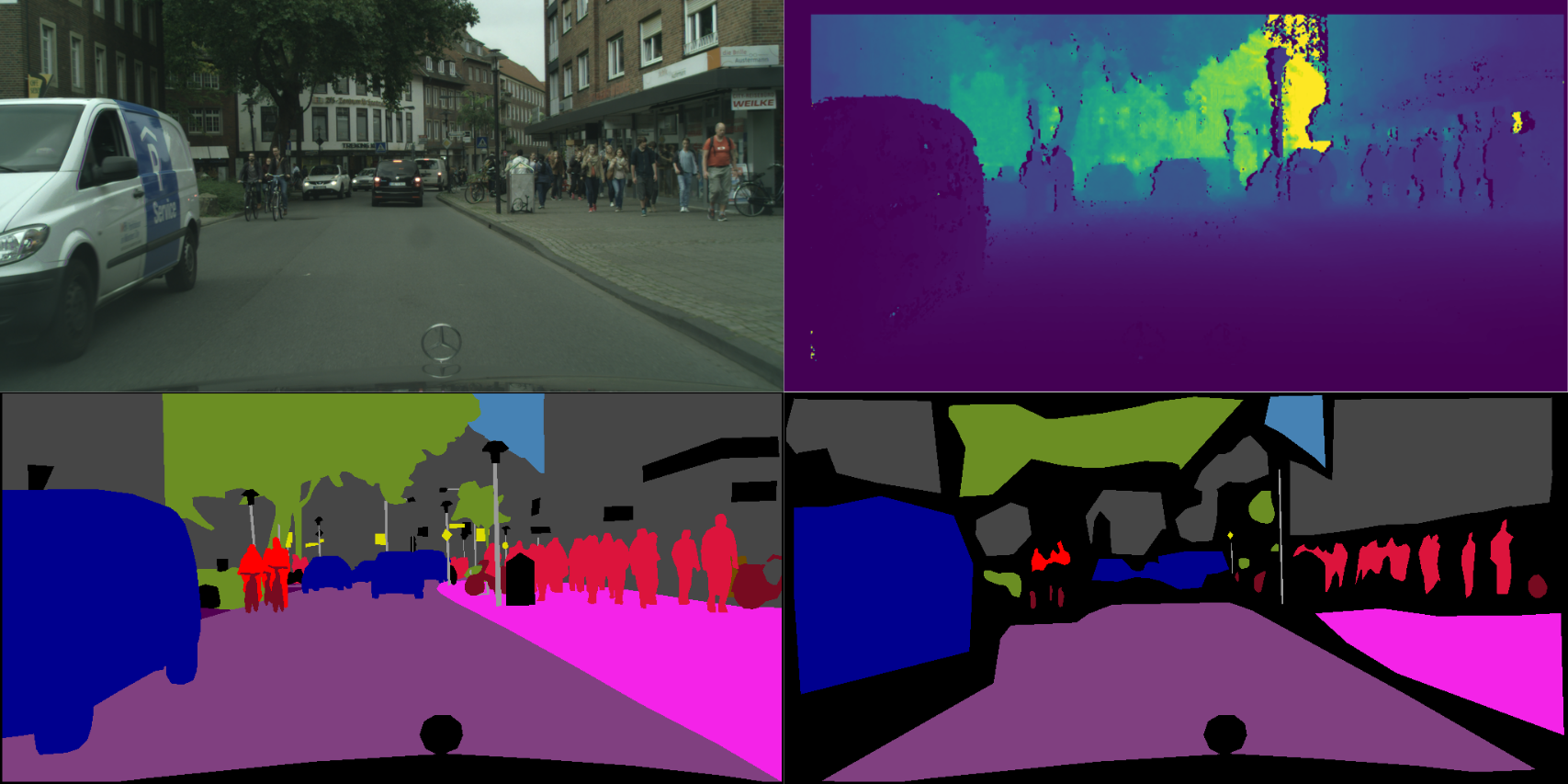}
   \caption{Annotations of Cityscapes dataset (original image, depth map, fine semantic map, coarse semantic map). While coarse annotations only outline the most prominent objects via polygons, pixel-exact fine annotations are regarded as the ground truth.}
\label{fig:coarse_vs_fine_appendix}
\end{figure}

From the Cityscapes semantic segmentation benchmark dataset (CITY) \cite{cordts16}, we use the 5000 images with fine semantic and disparity annotations, and the 19998 images with coarse semantic annotations (see \cref{fig:coarse_vs_fine_appendix}). We use the official splits for training and validation with 2975 and 500 samples, respectively. Since labels for the 1525 test samples are not publicly available, we report best performances achieved on the validation split. For JSRT \cite{shiraishi00}, a 80:10:10 split is applied, resulting in 199, 24, and 24 samples for training, validation, and testing, respectively. For CITY, we use 34 semantic classes with 8 categories for training and the official 19 classes for evaluation. For JSRT, we use 6 classes for training and evaluation: background, left lung, right lung, heart, left clavicle, and right clavicle. We rescale all samples to a resolution of 512$\times$1024 pixels for CITY and 512$\times$512 pixels for JSRT. For CITY disparity maps, we observe that some pixels near image borders are labeled incorrectly. We manually mask out these erroneous labels and treat them as invalid measurements. 

\subsection{Models, Methods, Training and Evaluation}
\label{sec:train_and_eval_appendix}
\textbf{Models:} We use a U-Net \cite{ronneberger15} variant with residual blocks \cite{he16} and instance normalization \cite{ulyanov16} as our default architecture, resulting in 9 levels (excluding the label-level $l=0$) with respective feature map resolutions of $[512\times1024, 256\times512, 128\times256, 64\times128, 32\times64, 16\times32, 8\times16, 4\times8, 2\times4]$ pixels. The respective number of feature maps are $[6$, $12$, $24$, $48$, $96$, $192$, $384$, $768$, $1536]$, resulting in a total of 73.15m learnable parameters.

For some experiments, we use a ConvNeXt Tiny with pre-trained weights for feature extraction \cite{liu22}. Pre-training was performed on ADE20K \cite{zhou17} for semantic segmentation. We use the publicly available, pre-trained weights provided by \cite{liu22}. The ConvNeXt model comprises 4 levels with feature map resolutions of $[128, 64, 32, 16]$ and corresponding number of feature maps $[96, 192, 384, 786]$. To be able to use it with our transduction module, we complete feature pyramids, i.e., add missing resolutions, via interpolation and extrapolation of neighboring representations. For ConvNeXt, we use the same number of feature maps at respective resolutions as for U-Net.

\textbf{Method:} For memory search, we set $n_{ch}=4^{dim}$ for all levels. For MP, we set $\lambda=1$, $k=16$. Convergence is typically achieved within 32 MP steps. For test-time augmentation (TTA), we use 3 forward passes with down-scaled and reflection-padded variants of $x_q$. Registration of down-scaled variants with the original variant is achieved by up-scaling and center-cropping the respective predictions. We down-scale inputs by factors $s=\{0.8, 0.9, 1\}$, where $s=1$ implies no transformation. A final prediction is obtained by averaging over all passes.

\textbf{Training:} Some of our experiments (baseline methods and pre-training for PARMESAN) require training of learnable parameters $\theta$. We use the cross-entropy loss as the default loss function for semantic segmentation. For monocular depth estimation, we regress $d:=\textit{log}\textit{(depth)}$ and use the scale-invariant loss \cite{eigen14} with $\lambda=0.5$. We mask out invalid labels both during training and evaluation. Training is performed via stochastic gradient descent \cite{robbins51}, utilizing the backpropagation algorithm \cite{kelley60,rumelhart86} and the AdamW optimizer \cite{kingma14,loshchilov18} with default hyperparameters, a minibatch size of 8, a constant learning rate of 0.0001, and a weight decay of 0.00001. We apply gradient-norm clipping \cite{pascanu13}, allowing maximum gradient norms of 50. We train our models on randomly cropped patches of 512$\times$512 pixels for CITY and on full images of 512$\times$512 pixels for JSRT. For data augmentation during training, we use horizontal flips ($p=0.5$, not for JSRT), vertical flips ($p=0.5$, not for JSRT), rotations ($p=0.5$, $\pm25^\circ$), Gaussian noise ($p=0.5$, $\sigma=0.05$), as well as changes in brightness, saturation, and contrast.

\textbf{Evaluation:} We use exponential moving average models \cite{polyak90,ruppert88,tarvainen17} with a rate of 0.99 for evaluation. Early stopping \cite{morgan89} with a patience of 5 and a validation verbosity of 3000 update steps is used (400 for JSRT). Early stopping in CL experiments only depends on the performance on available classes at the respective CL step. Since classes are highly imbalanced in the \textit{13-1 (7) cl} scenario, we apply early stopping using both old and current classes for JOINT and GDUMB.

We use the Intersection over Union (IoU) \cite{jaccard12,tanimoto58} as evaluation metric for semantic segmentation, which is defined as the ratio between the size of the intersection and the size of the union of two finite sample sets, measuring their similarity. In this work, we use IoU to measure predictive performance of a model for the task of pixel-wise, multi-class classification, i.e., semantic segmentation. We employ the standard evaluation procedure as defined in \cite{cordts16}. Specifically, we apply the mean IoU ($\miou{}$), where scores are calculated globally without favoring any class in particular. This variant is also referred to as ``micro''-averaging.

\begin{figure}[t]
	\centering
   \includegraphics[width=1.\linewidth]{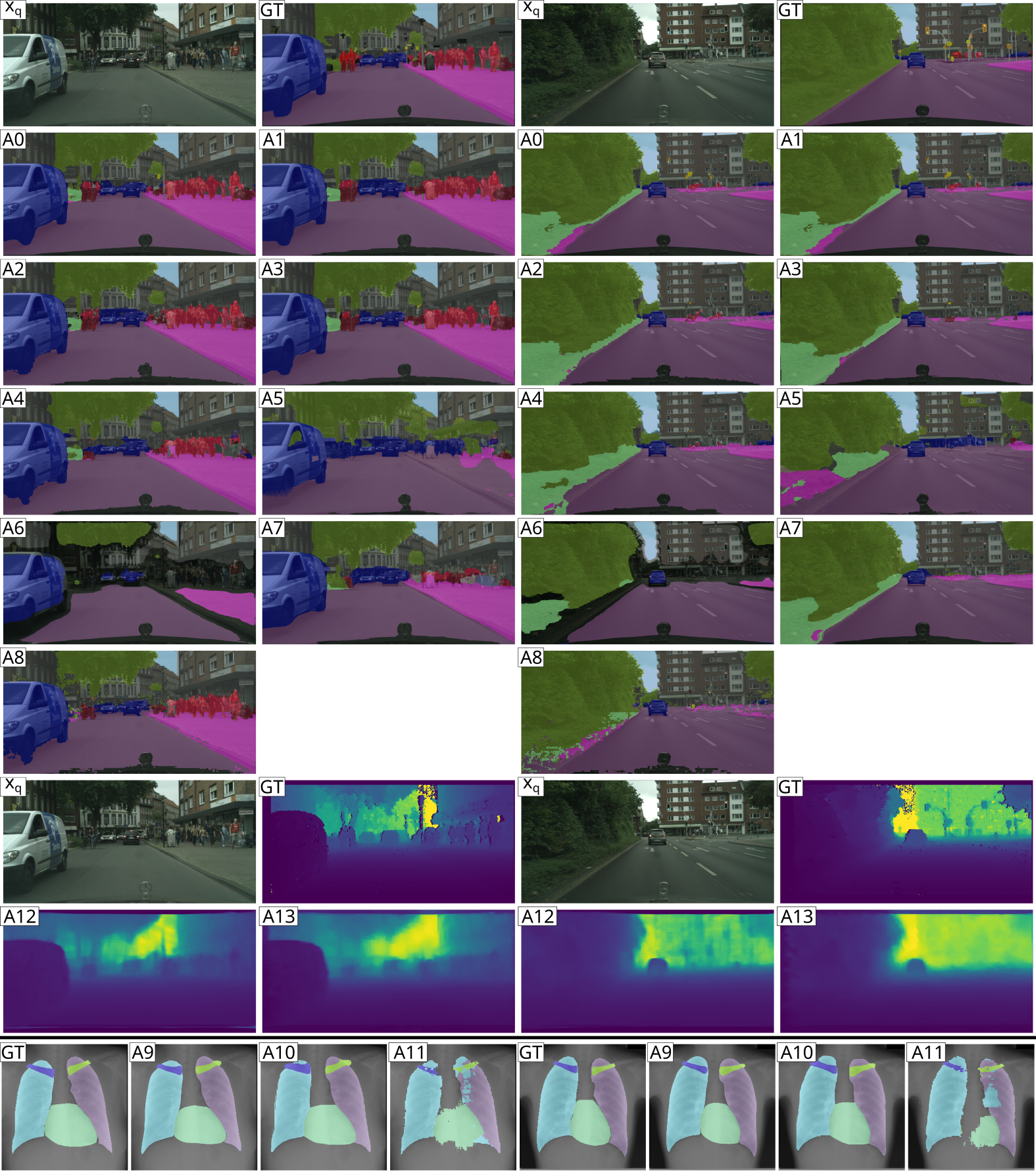}
   \caption{Model predictions for all ablation studies with two different query inputs.}
\label{fig:predictions_appendix_A}
\end{figure}

\begin{figure}[t]
	\centering
   \includegraphics[width=1.\linewidth]{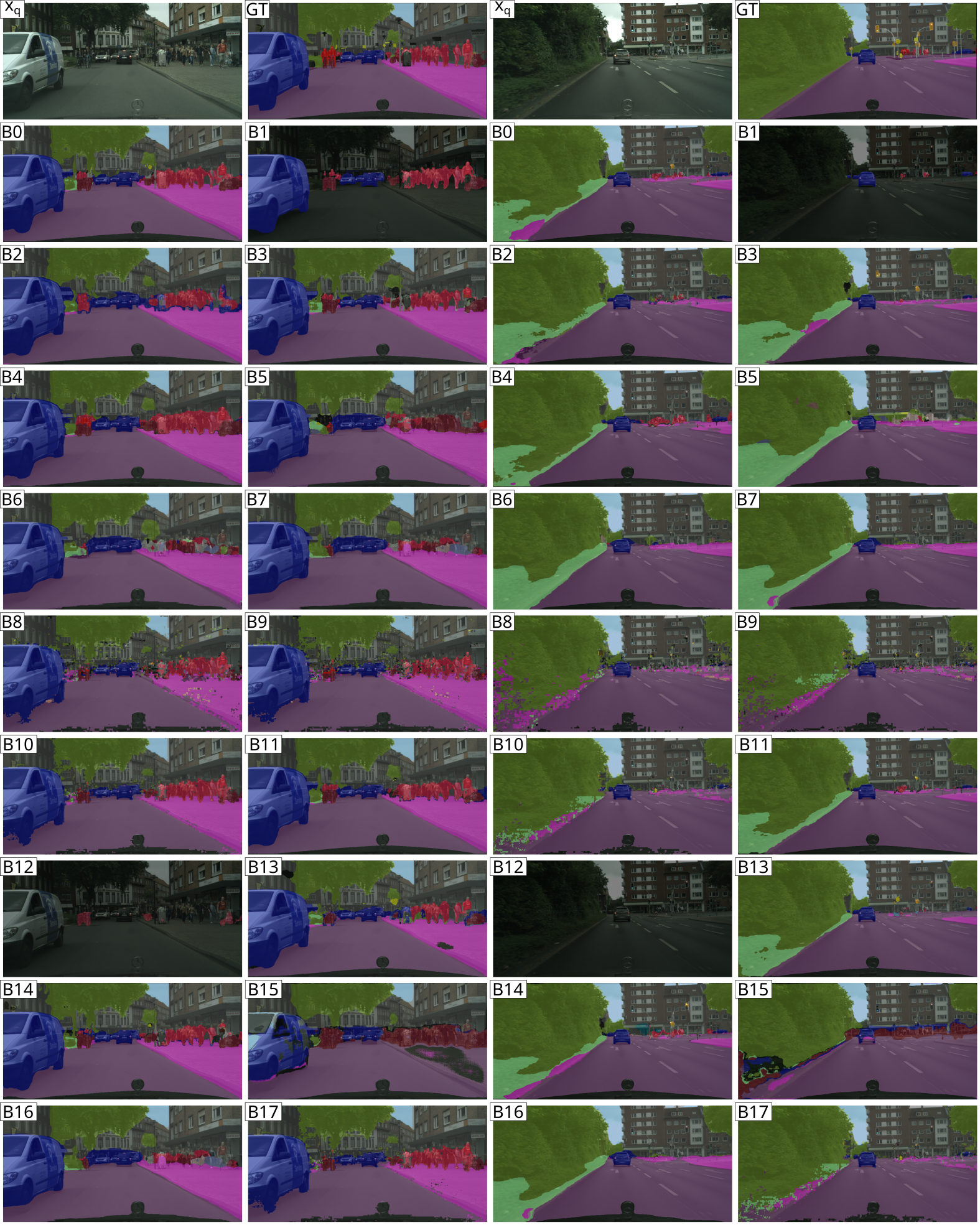}
   \caption{Model predictions for all CL experiments with two different query inputs.}
\label{fig:predictions_appendix_B}
\end{figure}

\textbf{Results:} We show qualitative prediction results for all ablation studies and CL experiments experiments in \cref{fig:predictions_appendix_A} and \cref{fig:predictions_appendix_B}, respectively. Experiments A14, A15, and A16 are not mentioned in the main paper text. Results in \cref{tab:results_efficiency} indicate moderate memory requirements for our method. Performance remains robust in case of low $m$ (A14) or when using less channels (A15, A16). Iterative MP and sequentially executed TTA lead to higher $\tau_i$ w.r.t. our best setup (A1 / A2 / A3).

\begin{table}[t]
	\scriptsize
	\centering
	\caption{Size of $M$ (samples stored on GPU), peak allocated GPU memory $\mathit{C}$ at inference, inference speed $\tau_i$, and predictive performance on CITY.}
	\begin{tabular}{l|l|l|l|l|l|l|l}
		\hline
		ID & Method \& Setup & $M$ [GB] & $\mathit{C}$ [GB] & $\tau_i$ [s] & $\tau_l$ [s] & $\miou{cl}$ & $\miou{cat}$ \\
		\hline
		A0 & JOINT & \none & 0.93 & 0.0068 & 3.72 & 82.5 & 87.4 \\
		A1 & ours & 6.40 & 11.0 & 1.43 & 0.012 & 82.1 & 87.0\\
		A2 & ours, no MP & 6.40 & 10.9 & 0.25 & 0.012 & 81.4 & 86.4\\
		A3 & ours, no TTA & 6.40 & 10.8 & 0.48 & 0.012 & 81.2 & 86.3\\
		A14 & ours, $m \smalleq 4, \phi \smalleq 1$ & 0.009 & 4.44 & 1.25 & 0.011 & 72.8 & 78.1\\
		A15 & ours, 75\% channels & 4.82 & 9.10 & 1.37 & 0.011 & 82.0 & 86.9\\ 
		A16 & ours, 50\% channels & 3.29 & 7.40 & 1.32 & 0.010 & 81.8 & 86.7\\ 
		\hline
	\end{tabular}
	\label{tab:results_efficiency}
\end{table}

\subsection{Continual Learning Setup}
\label{sec:CL_appendix}
The goal in CL is learning a sequence of $N$ tasks $\mathcal{T}=((T_1, \textit{D}_1), ..., (T_N, \textit{D}_N))$ on respective datasets $\mathcal{D} = (\textit{D}_1, ..., \textit{D}_N)$. Old data $\textit{D}_j$ is inaccessible for CL step $i$ for $j<i$. For memory-based CL methods, data from $\textit{D}_j$ can still be retained if it is stored in $M$. Given a task-solving model (''solver'') $S: X\rightarrow Z$ with inputs $x \in X$, predictions $y \in Z$, and labels $z \in Z$, the objective is minimizing the unbiased sum of losses $\mathbb{E}_{x, z} [L(y, z)]_\mathcal{T}$ over all tasks on the entire data distribution.

We employ two different CL scenarios: \textit{2-2 (4) cat} and \textit{13-1 (7) cl}. At \textit{2-2 (4) cat}, data at every CL step comprises labels for all classes of different categories: $\textit{D}_1$: \textit{void} \& \textit{flat}, $\textit{D}_2$: \textit{construction} \& \textit{object}, $\textit{D}_3$: \textit{nature} \& \textit{sky}, and $\textit{D}_4$: \textit{human} \& \textit{vehicle}. As proposed by Yang et al.~\cite{yang23}, the \textit{13-1 (7) cl} scenario starts with 13 classes at CL step $i=1$ (i.e., \textit{road, sidewalk, building, wall, fence, pole, traffic light, traffic sign, vegetation, terrain, sky, person, rider}) and adds one new class at every following CL step (i.e., \textit{car, truck, bus, train, motorcycle, bicycle}).

In contrast to image classification, samples for semantic segmentation can contain classes that belong to different CL steps, which is referred to as overlapping labels. For the CI-CL scenario, this requires introducing a dedicated background class to avoid data-leakage between different CL steps. Therefore, non-available classes are re-labeled as background in each CL step. This implies that the background class is different for different CL steps. As proposed in recent work \cite{cermelli20}, we ignore background pixels for validation.

Our method supports any combination of the three CL scenarios TI, DI, and CI. Although we focus on studying the offline CL scenario in this paper, our method can also be used for online CL, where one does not have access to the entire dataset of a specific task in order to learn continually.

In order to keep labels consistent for REPLAY, we employ an adaptive masking approach. For samples of the current CL step $i$, we mask out all background pixels, i.e., pixels which do not belong to currently active classes. In contrast, for memory samples we mask out all pixels which belong to currently active classes, i.e., keeping pixels of old classes and the background. Training minibatches for REPLAY consist of standard samples in addition to replay samples, which we select randomly from $M$. For REPLAY, we observe that using 4 or more replay samples from $M$ per update step leads to strong memory overfitting, which in turn leads to reduced validation performance. For $m=940$, we achieve best results with minibatches composed of 3 replay samples and 8 standard samples.

For MiB and REPLAY baselines, the background class is learned in the first CL step. During successive CL steps, background is partially unlearned again depending on new classes.

\end{document}